\documentclass[11pt,letterpaper]{article}
\usepackage{cogsys}
\usepackage[T1]{fontenc}
\usepackage{times}
\usepackage{amsmath}
\usepackage[pdftex]{graphicx} 
\usepackage{multirow}
\usepackage{bm}
\usepackage{hang}
\usepackage{natbib}
\usepackage{tablefootnote}
\usepackage[flushleft]{threeparttable}

\usepackage[labelsep=period]{caption}
\usepackage[labelformat=simple]{subcaption}

\newcommand\myeq{\mkern1.5mu{=}\mkern1.5mu}

\usepackage[]{hyperref}
\hypersetup{
    pdftitle={Visual-Imagery-Based Analogical Construction in Geometric Matrix Reasoning Task},
    pdfauthor={Yuan Yang, Maithilee Kunda, and Keith McGreggor},
    bookmarksnumbered=true,     
    bookmarksopen=true,         
    bookmarksopenlevel=1,       
    colorlinks=true,
    citecolor=[rgb]{0, 0.5, 0.5},
    linkcolor=[rgb]{0, 0.5, 0.5},
    pdfstartview=FitH,          
    pdfpagemode=UseOutlines,    
    pdfpagelayout=SinglePage
}

\DeclareMathOperator*{\argmax}{arg\,max}

\cogsysheading{X}{20XX}{1-6}{X/20XX}{X/20XX}

\ShortHeadings{Visual-Imagery-Based Analogical Construction}
              {Y.\ Yang, K.\ McGreggor, and M.\ Kunda}

\graphicspath{{pic/}}

\begin{document} 

\title{Visual-Imagery-Based Analogical Construction\\ in Geometric Matrix Reasoning Task}
 
\author{Yuan Yang}{yuan.yang@vanderbilt.edu}
\address{Electrical Engineering and Computer Science, Vanderbilt University, Nashville, TN 37235 USA}
\author{Keith McGreggor}{keith.mcgreggor@gatech.edu}
\address{School of Interactive Computing, Georgia Tech, Atlanta, GA 30308 USA}
\author{Maithilee Kunda}{mkunda@vanderbilt.edu}
\address{Electrical Engineering and Computer Science, Vanderbilt University, Nashville, TN 37235 USA}

\vskip 0.2in

 
\begin{abstract}
Analogical reasoning fundamentally involves exploiting redundancy in a given task, but there are various strategies for an intelligent agent to identify and exploit such redundancy, often resulting in very different levels of reasoning ability.  We explore such variations of analogy in geometric reasoning task, namely the Raven's Progressive Matrices. We show how different analogical constructions used by the same basic imagery-based computational model --- varying only in how they ``slice'' a matrix problem into parts and search within/across these parts --- achieve very different test scores, substantially overlapping the range of human performance. Our findings suggest that the ability to build effective high-level analogical constructions is as important as competencies in low-level reasoning, which raises interesting questions about the extent to which building the ``right'' analogies contributes to individual differences in human reasoning and how intelligent agents might learn to build among different constructions in the first place.
\end{abstract}

\section{Introduction}

\begin{figure}[t]
    \centering
    \begin{subfigure}{0.44\textwidth}
        \centering
        \includegraphics[width=\textwidth]{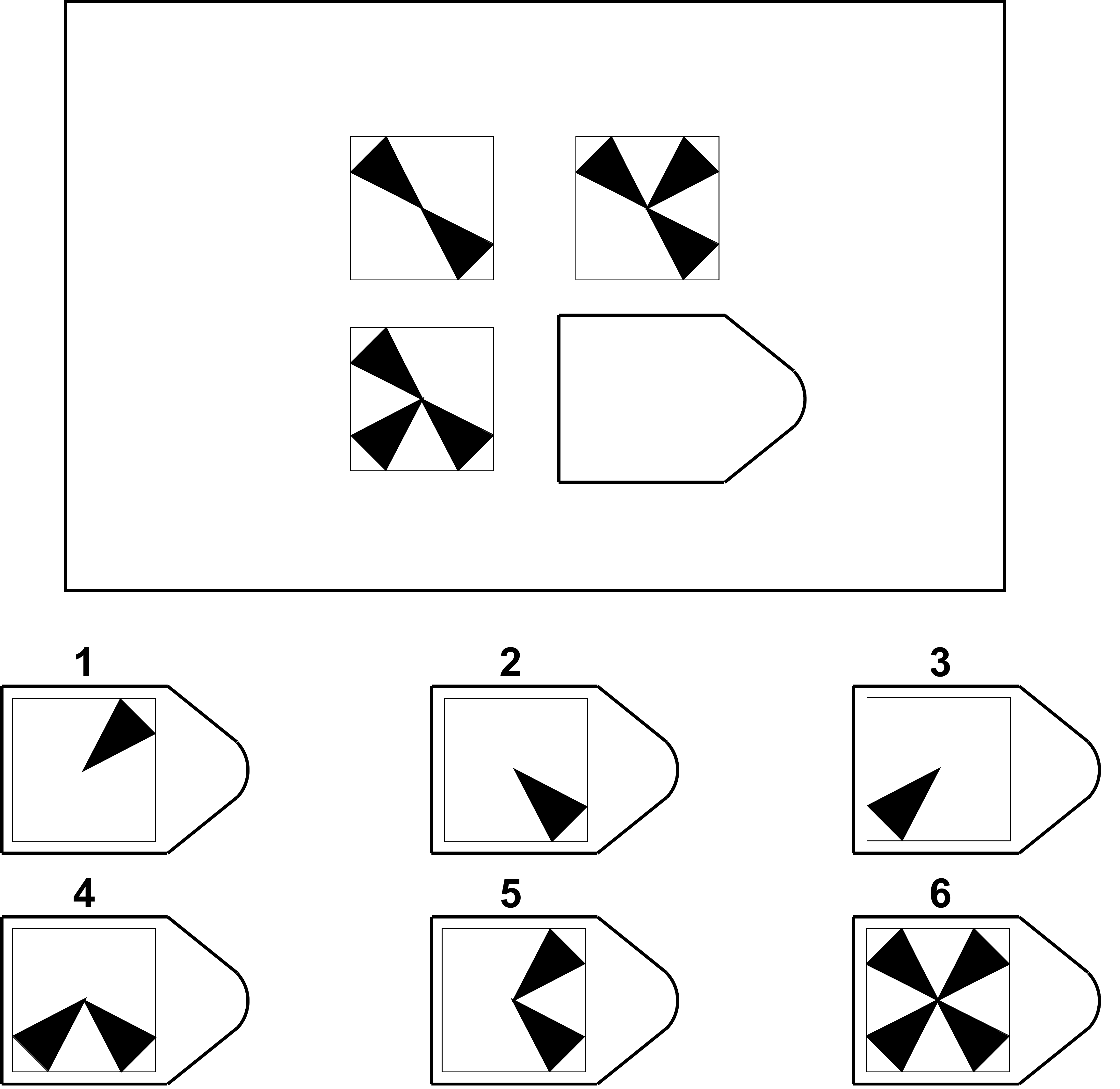} 
        \caption{2$\times$2 RPM-like problem}
        \label{fig:example_RESW_a}
    \end{subfigure}
    \hspace{0.05\textwidth}
    \begin{subfigure}{0.45\textwidth}
        \centering
        \includegraphics[width=.97\textwidth]{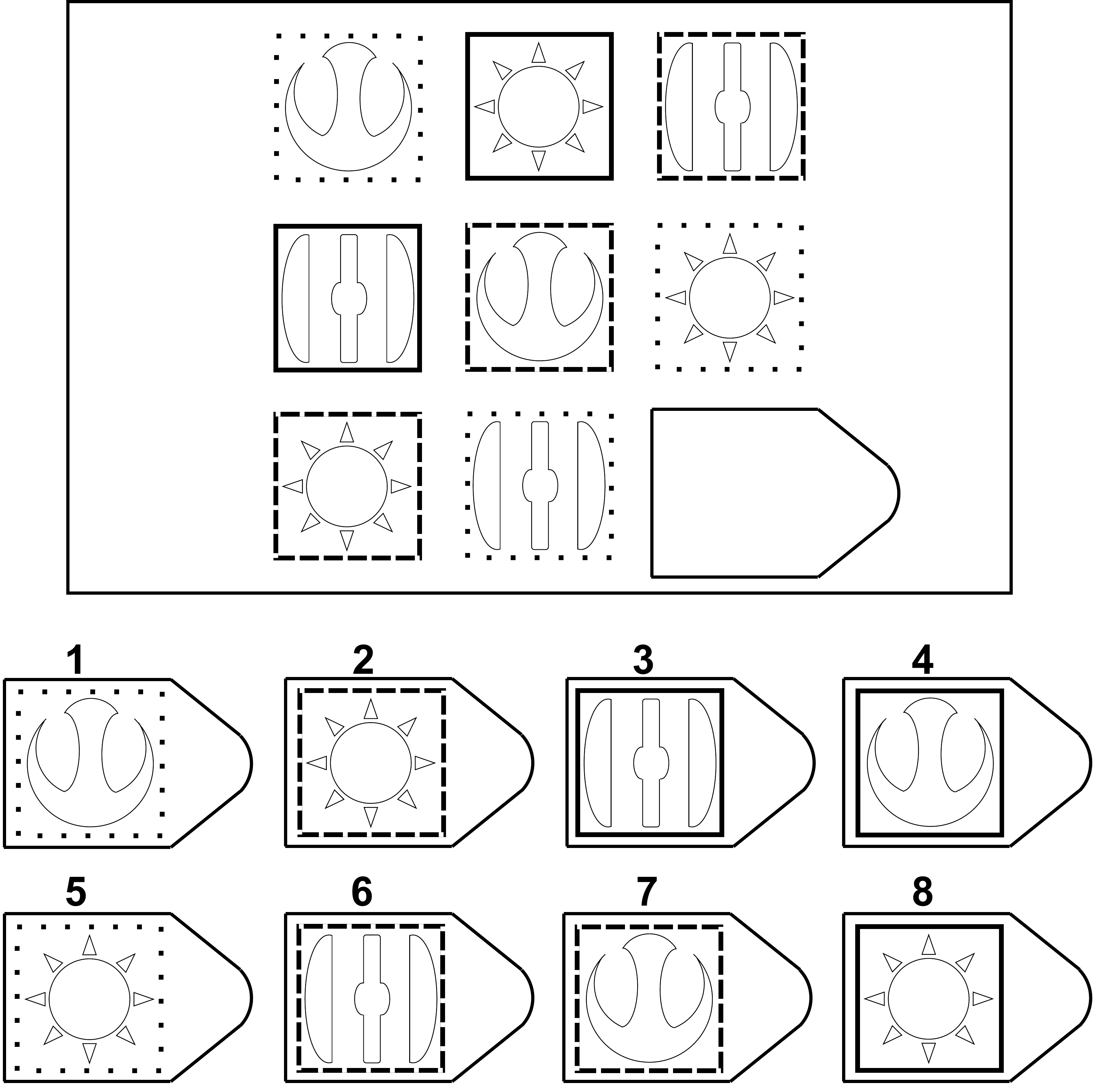}
        \caption{3$\times$3 RPM-like problem}
        \label{fig:example_RESW_b}
    \end{subfigure}
    \caption{Example problems: Real RPM problems are not shown to protect the secrecy of the test.}
    \label{fig:example_RESW}
\end{figure}


Raven's Progressive Matrices (RPM) is a widely used intelligence test that contains geometric reasoning problems like those shown in Figure \ref{fig:example_RESW} --- including 2$\times$2 problems (left) and 3$\times$3 problems (right).  The task is to select an answer from the options provided below the matrix that best completes it so that the relations in parallel rows and columns (and diagonals in some cases) form meaningful analogies.

How do you solve such problems?  Your solution process is likely to involve constructing analogies from the problem elements --- one row or column becomes the source, another row or column becomes the target, you find a mapping between them, and finally you transfer information from the source to the target to produce an answer --- but there are many possible ways to construct analogies.  For the 2$\times$2 problem on the left, you might construct analogies based on rows or columns.  For the 3$\times$3 problem on the right, there are far more variations.  Perhaps you just focus on the top and bottom rows, ignoring the middle row completely.  Or maybe you look at the top row first, use the second row to ``verify'' your hypothesis, and then try to fill in the bottom row.

When taking the RPM test, no one tells you how to construct these various analogies to get to the answer.  Some research suggests that the ability to construct abstract analogical relations is an innate capacity that distinguishes humans from other species \citep{hespos2020structure}. The RPM was specifically designed to test a person's \textit{eductive ability} to extract information from and make sense of a complex situation \citep{raven1998manual}, where analogies are often indispensable.  Previous computational models have explored many different dimensions of matrix reasoning, including the capacity for subgoaling \citep{carpenter1990one}, pattern matching \citep{cirillo2010anthropomorphic}, rule induction \citep{rasmussen2011neural}, and dynamically re-representing and re-organizing visual elements \citep{lovett2017modeling}.  

In this paper, we present a systematic examination of another dimension of matrix reasoning: how one constructs analogies from matrix elements.  As our base model, we use the Affine and Set Transformation Induction (ASTI) model, which operates on scanned, pixel-based images from the RPM test booklet and uses affine transformations and set operations to reason about image differences \citep{kunda2013computational, kunda2013visual}. Our contributions include:
\cbullet A three-level hierarchy for solving RPM problems.  First, at the level of images, one can search across a set of image \textbf{transformations} to interpret relationships within a given pair or triple of images (e.g., to explain the variation across a row, column, or diagonal).  Second, at the level of a problem matrix, one can search across different \textbf{analogies} to find transfers of relationships across different pairs or triples of images.
Third, at the highest level, one can use alternative \textbf{integration strategies} that specify how to combine results from different levels to produce the final answer.

\cbullet A finer taxonomy of the \textbf{option-usage strategy} for solving RPM problems, which are traditionally categorized into constructive matching and response elimination \citep{bethell1984adaptive}. We further divide constructive matching into \textbf{option-free} and \textbf{option-informed} constructive matchings, which describe usages in practice more precisely.

\cbullet A demonstration that a certain combination of transformations, analogies, and integration strategy solves 57/60 problems on the Raven's Standard Progressive Matrices test, which shows that these representations and inference mechanisms are expressive and effective for building computational solutions to this type of tasks.

\cbullet Systematic ablation experiments that show test performance varies widely as a function of overall analogy constructions and that it covers almost the entire range of human performance reported in the studies of the RPM test.

\section{Description of the ASTI+ Model}

In this section, we present the ASTI+ model for solving RPM problems. We provide detailed and formal description of its core dimensions: problem representations, similarity metrics, image transformations, matrix analogies, and integration strategies. Based on the description, we then compare the ASTI+ model to its predecessor --- the ASTI model \citep{kunda2013computational,kunda2013visual}.

\subsection{Problem Representations}

\begin{figure}[t]
    \centering
    \includegraphics[width=0.8\textwidth]{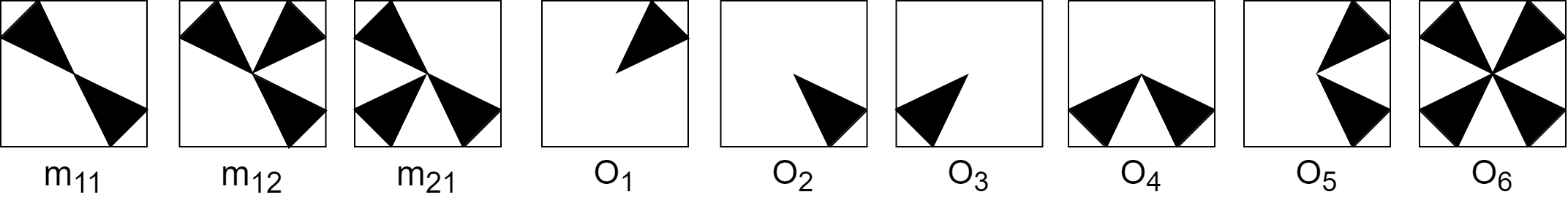}
    \caption{Illustration of input to our model for the 2$\times$2 example problem:  $\mathsf{m_{ij}}$ is the matrix entry in Row $\mathsf{i}$ and Column $\mathsf{j}$, and $\mathsf{O_k}$ is the $\mathsf{k}$-th answer option.}
    \label{fig:input_RE}
\end{figure}

Since the standard RPM is in black and white, we represent each problem as a binary (i.e. pure black and white) image. Note that this is equivalent to representing an image as a set of black pixels with each pixel identified by its coordinates in the image. Throughout this paper, we use these two representations interchangeably. Binary images are generated from grayscale scannings of RPM test booklet and we select a threshold manually to convert grayscale values to binary values. We use an RPM-specific automated image-processing pipeline \citep{kunda2013visual} to decompose each full test page into images of individual matrix entries and answer options, as shown in Figure \ref{fig:input_RE}. We then feed these individual images as inputs to the ASTI+ model.

\subsection{Similarity Metrics}
\label{sec:sim_metrics}
The first element of our model specifies how to measure similarity between images. For this purpose, we use the Jaccard index and the asymmetric Jaccard index, as shown in Equation \eqref{eq:jaccard} and \eqref{eq:asymmetric_jaccard}:
\begin{align}
    J(A,B) &=\frac{|A \cap B|}{|A \cup B|} \label{eq:jaccard} \\
    J_A(A,B) &=\frac{|A \cap B|}{|A|} \label{eq:asymmetric_jaccard}
\end{align}
where $A$ and $B$ are two sets representing two binary images. Equation \eqref{eq:asymmetric_jaccard} is asymmetric because $J_A(A,B) \neq J_A(B,A)$ and measures the extent to which $A$ is inside (or a subset of) $B$.

A problem with Equation \eqref{eq:jaccard} and \eqref{eq:asymmetric_jaccard} is that $A$ and $B$ should be properly aligned. That is, $A$ and $B$ should have the same shape and size, and pixels in $A$ and $B$ belonging to the same object should have the same coordinates in $A$ and $B$. However, the images of matrix entries and options come in various shapes and sizes. We take a simple but robust approach to this problem --- slide one image over the other, calculate a similarity value at every relative position, and select the maximum. In the process of sliding, images are padded to have the same shape and size.
\begin{align}
    S(A,B) &= (J(A,B),pos_{AB})  \label{eq:jaccard_sliding} \\
    S_A(A,B) &= (J_A(A,B),pos_{AB}, pos_{DA}, D) \label{eq:asymmetric_jaccard_sliding}
\end{align}

As a result, similarity calculation in our model is defined by Equation \eqref{eq:jaccard_sliding} and \eqref{eq:asymmetric_jaccard_sliding}, where $J(A,B)$ and $J_A(A,B)$ are the maximum\footnote{When the maximum is achieved at multiple relative positions, we take the least shifted one. If multiple such least shifted positions exist, then the image must contain some symmetric structure. In this case, all of them are equally representative and we only need to select one in a consistent manner, for example, always select the first one. Of course, there exist other methods to resolve this issue.} similarity values at the relative position $pos_{AB}$ of $A$ to $B$. In Equation \eqref{eq:asymmetric_jaccard_sliding}, $D=B-A$ is the difference between $A$ and $B$ when the maximum is reached, and $pos_{DA}$ is the relative position of $D$ to $A$.

\subsection{Transformations}
\label{sec:trans}
The second element specifies low-level visuospatial knowledge about the domain. ASTI+ represents this content as a discrete set of image transformations that map from one or more input images to an output image.  These functions operate on images at the pixel level, without re-representing visual information in terms of higher-order features.  Although these functions were defined manually, based largely on inspections of the Raven's test, important directions for future work include expanding them to include higher-order features and concepts, as well as learning from perceptual experience \citep{michelson2019learning}.

\begin{figure}[t]
    \centering
    \includegraphics[width=\textwidth]{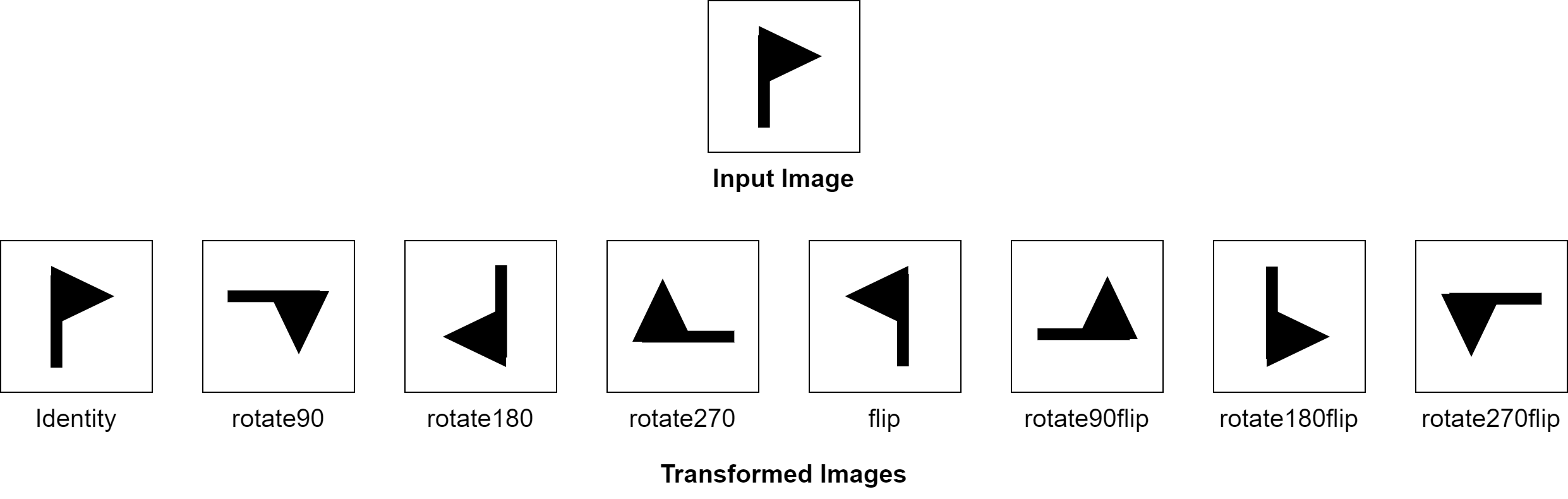}
    \caption{Illustrations of affine transformations used in our model.}
    \label{fig:affine_trans}
\end{figure}

\begin{figure}[t]
    \begin{subfigure}[t]{0.5\textwidth}
        \captionsetup{singlelinecheck=false}
        \centering
        \caption{}
        \vskip -0.12in
        \includegraphics[width=\textwidth]{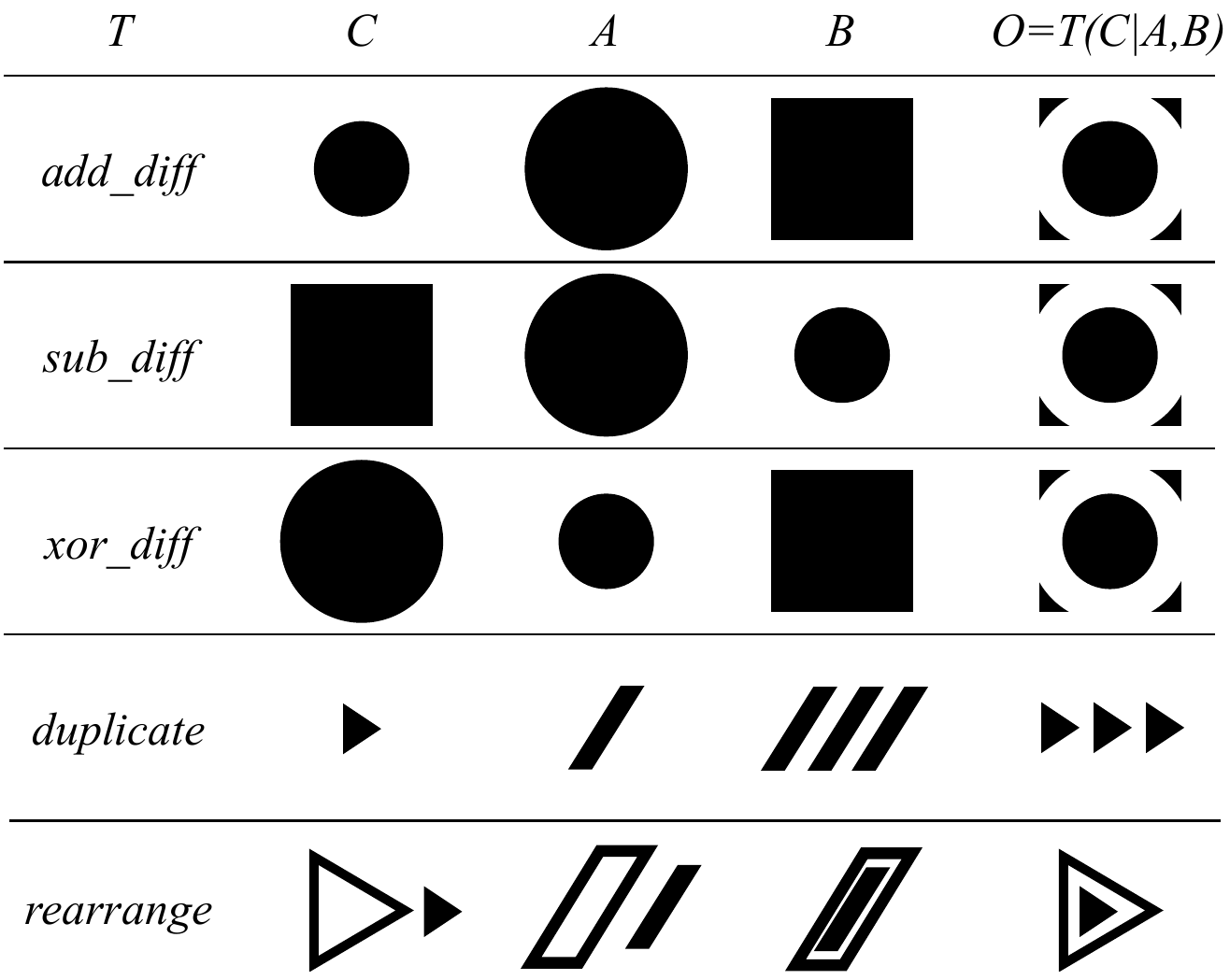} 
        \label{fig:set_unary}
    \end{subfigure}
    \begin{subfigure}[t]{0.5\textwidth}
        \captionsetup{singlelinecheck=false}
        \centering
        \caption{}
        \vskip -0.12in
        \includegraphics[width=\textwidth]{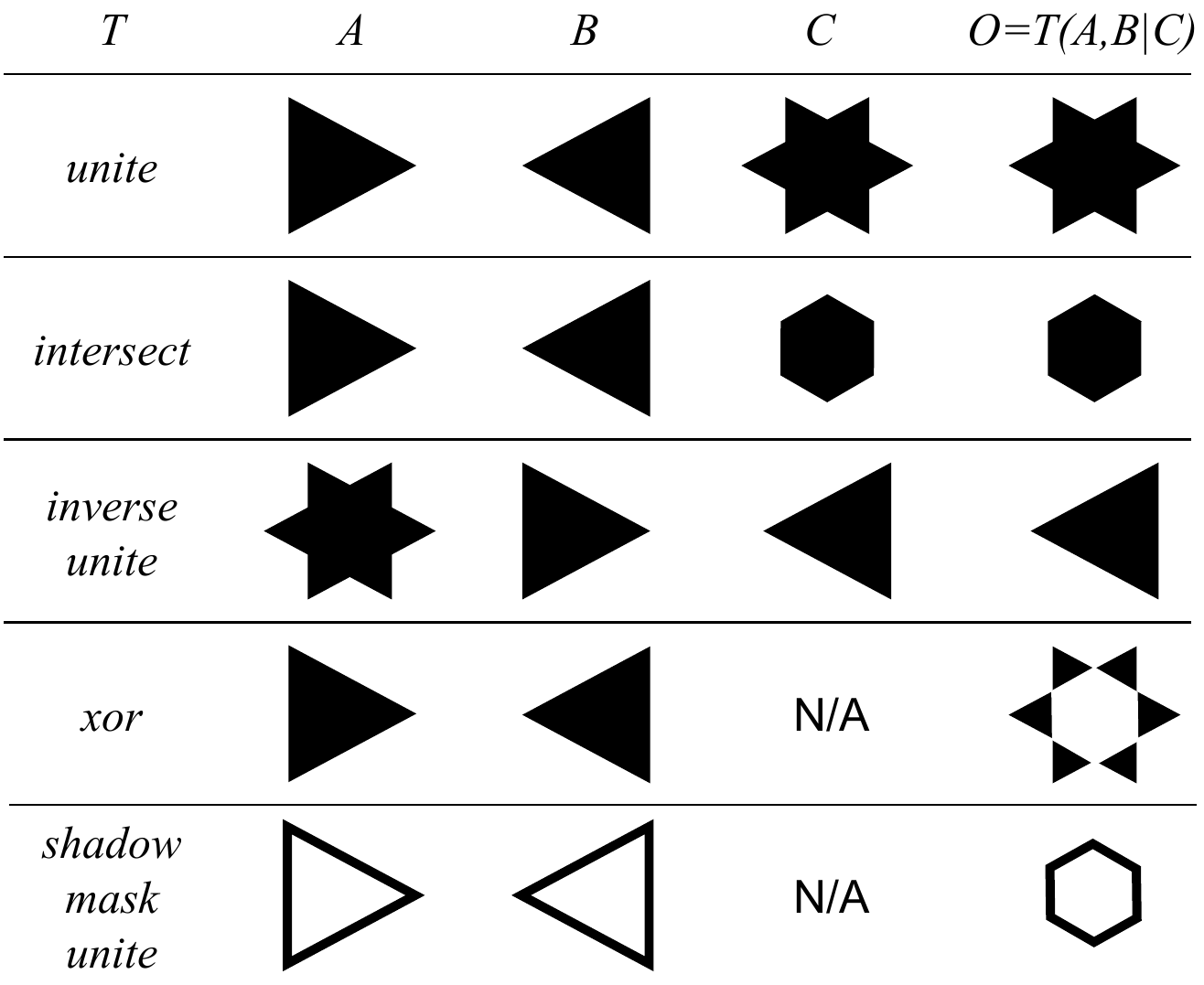}
        \label{fig:set_binary}
    \end{subfigure}
    \caption{Illustrations of set transformations used in our model: (a) Given an analogy \textit{A:B::C:?} and an unary set transformation $T$, the output image is $O=T(C|A,B)$, where $C$ is the input, and $B$ and $C$ are parameters of $T$; (b) Given an analogy \textit{A:B:C::D:E:?} and a binary set transformation $T$, the output image is $O=T(A,B|C)$ when $T$ is applied on \textit{A:B:C}, where $A$ and $B$ are the inputs, and $C$ is a parameter of $T$, or $O=T(D,E|O')$ when $T$ is applied on \textit{D:E:?}, where $O'$ is an option of the RPM problem.}
    \label{fig:set_trans}
\end{figure}

ASTI+ includes two types of image transformations: unary and binary, which take a single input image and two input images, respectively. All ASTI+ transformations are based on fundamental affine transformations and set operations. These extend the original collections proposed in earlier ASTI research \citep{kunda2013computational,kunda2013visual}. ASTI+ includes nine unary affine transformations: eight rectilinear rotations/reflections, as shown in Figure \ref{fig:affine_trans}, and a ninth scaling transformation that doubles the area of the input image. There are also 11 additional set transformations: five unary and five binary, as shown in Figure \ref{fig:set_unary} and \ref{fig:set_binary} respectively, and one hybrid unary/binary transformation. Table \ref{table:transformations} gives details of each transformation.  Unary transformations are defined relative to analogies between pairs of images, such as \textit{A:B::C:D} for images \textit{A}, \textit{B}, \textit{C} and \textit{D}.  Binary transformations are defined relative to analogies between trios of images, such as \textit{A:B:C::D:E:F} for images \textit{A}, \textit{B}, \textit{C}, \textit{D}, \textit{E} and \textit{F}. 

\begin{table}[ht]
    \centering
    \caption{Details of unary, binary, and hybrid unary/binary transformations.}
    \label{table:transformations}
\begin{threeparttable}
    \textsf{
        \vspace{-6pt}
        \rule{\textwidth}{0.4pt}
        \vspace{-15pt}
        \scriptsize{
            \begin{compactlabel}{$PSD(D, E|A,B,C)^*$}
                \item[$add\_\mathit{diff}(C|A,B)$] Calculate $S_A(A,B)=(\cdots, pos_{DA}, D)$. Align $C$ and $D$ using $pos_{DA|A=C}$.  Output $O=C \cup D$.
                \item[$sub\_\mathit{diff}(C|A,B)$] Calculate $S_A(B,A)=(\cdots, pos_{BA}, pos_{DB}, D)$. Align $C$ and $D$ using $pos_{BA|A=C}$ and $pos_{DB}$.  Output  $O=C - D$.
                \item[$xor\_\mathit{diff}(C|A,B)$] Calculate $S(A,B)=(\cdots,pos_{AB})$. Align $A$ and $B$ using $pos_{AB}$, and calculate $D=A \oplus B$ and $pos_{DA}$. Align $C$ and $D$ using $pos_{DA|A=C}$. Output $O=C \oplus D$.
                \item[$duplicate(C|A,B)$] Let $O$ be an empty image of the same size as $B$. Calculate $S_A(A,B)=(\cdots,pos_{AB}, \cdots)$ and $B=B-A$ aligned by $pos_{AB}$, and copy $C$ to the position of $pos_{AB|A=C}$ in $O$. Repeat this until nothing is left in $B$. Output $O$.
                \item[$rearrange(C|A,B)$] Let $O$ be an empty image of the same size as $B$. Decompose $C$, $A$ and $B$ into connected components $C_1, C_2, \cdots, C_l$, $A_1, A_2, \cdots , A_m$ and $C_1, C_2, \cdots , C_n$. If $l=m=n$ is false, output a value indicating failure. Otherwise, find a permutation $f$ of $\{1, 2, \cdots , n\}$ that maximizes $\sum_{i=1}^{n} J(A_i, B_{f(i)})$ by calculating $S(A_i, B_j)=(J(A_i,B_j),pos_{A_iB_j})$ for each $i$ and each $j$. Find another permutation $g$ of $\{1, 2, \cdots , n\}$ that minimizes $\sum_{i=1}^{n} distance(C_i,A_{g(i)})$. Generate $O$ by copying $C_i$ to position of $pos_{A_{g(i)}B_{f(g(i))}|A_{g(i)}=C_i}$ in $O$ for all $i$. 
            \end{compactlabel}
        }
        \vspace{-12pt}
        \rule{\textwidth}{0.4pt}
        \vspace{-20pt}
        \scriptsize{
            \begin{compactlabel}{$PSD(D, E|A,B,C)^*$}
                \item[$unite(A,B|C)$] Calculate $S_A(A,C)=(\cdots,pos_{AC}, \cdots)$ and $S_A(B,C)=(\cdots,pos_{BC}, \cdots)$. Align $A$ and $B$ with $pos_{AC}$ and $poc_{BC}$. Output $O=A \cup B$.
                \item[$intersect(A,B|C)$] Calculate $S_A(C,A)=(\cdots,pos_{CA}, \cdots)$ and $S_A(C,B)=(\cdots,pos_{CB}, \cdots)$. Align $A$ and $B$ with $pos_{CA}$ and $poc_{CB}$. Output $O=A \cap B$.
                \item[$IU(A,B|C)$\tnote{*}] Calculate $S_A(B,A)=(\cdots,pos_{BA}, \cdots)$ and $S_A(C,A)=(\cdots,pos_{CA}, \cdots)$. Align $A$, $B$ and $C$ using $pos_{BA}$ and $pos_{CA}$.  Output image $O=A - (B - C)$.
                \item[$xor(A, B)$] Calculate $S(A,B)=(\cdots,pos_{AB})$. Align $A$ and $B$ by $pos_{AB}$. Output $O= A \oplus B$. 
                \item[$SMU(A, B)$\tnote{*}] Let $X$ and $Y$ be the shadows  of $A$ and $B$, where ``shadow'' is defined to be a copy of an image where any white area surrounded by black in the original image is colored black. Calculate $S(X, Y)=(\cdots,pos_{XY})$. Align $X$ and $Y$ using $pos_{XY}$, and calculate $M=X \cap Y$. Align $A$ and $B$ using $pos_{XY|X=A,Y=B}$. Output $O=M \cap (A \cup B)$.
            \end{compactlabel}
        }
        \vspace{-13pt}
        \rule{\textwidth}{0.4pt}
        \vspace{-20pt}
        \scriptsize{
            \begin{compactlabel}{$PSD(D, E|A,B,C)^*$}
                \item[$PSD(D, E|A,B,C)$\tnote{*}] Given analogy $A{:}B{:}C{::}D{:}E{:}?$, $preserving\_sub\_\mathit{diff}$ works as $sub\_\mathit{diff}(E|B,C)$. But it requires that $A \subset B \cap C$ and $D \subset E \cap O$, where $O$ is an option. Otherwise, output a value indicating failure. (This transformation is NOT shown in Figure \ref{fig:set_trans}.)
            \end{compactlabel}
        }
        \vspace{-15pt}
        \rule{\textwidth}{0.4pt}
        \vspace{-11pt}
    }
    \scriptsize{
        \begin{tablenotes}
            \item[*] $IU=inverse\_unite$, $SMU=shadow\_mask\_unite$, $PSD=preserving\_sub\_\mathit{diff}$
        \end{tablenotes}
    }
\end{threeparttable}
\end{table}

\subsection{Analogies}
\label{sec:anlgs}
The third element specifies how analogies are defined within a given RPM problem next. ASTI+ posits that an RPM analogy is composed of relations between matrix entries and that all parallel relations should be instantiated by the same transformation.  This assumption seems adequate for most problems on the Standard Raven's test, but items on the Advanced test or other geometric analogy tests may require considering multiple transformations \citep{carpenter1990one,kunda2015computational}.  

\begin{figure}
    \centering
    \begin{subfigure}{0.15\textwidth}
        \centering
        \includegraphics[width=\textwidth]{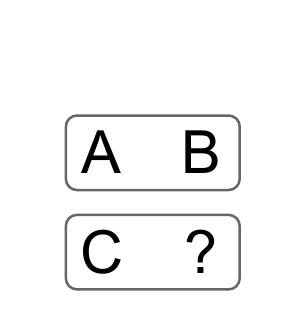} 
        \caption{}
        \label{fig:simple_anlgs_h22}
    \end{subfigure}
    \begin{subfigure}{0.15\textwidth}
        \centering
        \includegraphics[width=\textwidth]{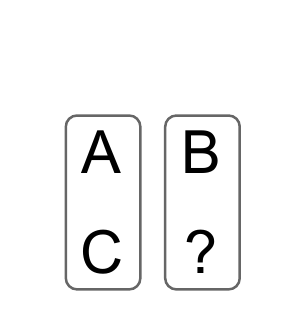} 
        \caption{}
        \label{fig:simple_anlgs_v22}
    \end{subfigure}
    \begin{subfigure}{0.15\textwidth}
        \centering
        \includegraphics[width=\textwidth]{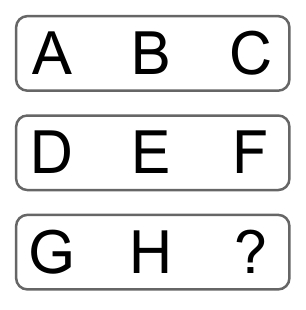} 
        \caption{}
        \label{fig:simple_anlgs_h33}
    \end{subfigure}
    \begin{subfigure}{0.15\textwidth}
        \centering
        \includegraphics[width=\textwidth]{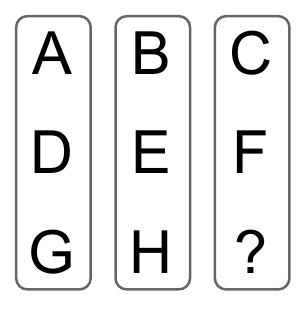} 
        \caption{}
        \label{fig:simple_anlgs_v33}
    \end{subfigure}
    \caption{Illustrations of simple analogies in RPM problems. Simple analogies reflect how a matrix layout is naturally perceived as rows or columns. Particularly, given 2$\times$2 matrices in (a) and (b), the row analogy is \textit{A:B::C:?} and the column analogy is \textit{A:C::B:?}; similarly, given 3$\times$3 matrix in (c) and (d), the row analogies include \textit{A:B:C::G:H:?} and \textit{D:E:F::G:H:?}, and the column analogies include \textit{A:D:G::C:F:?} and \textit{B:E:H::C:F:?}.}
    \label{fig:simple_anlgs}
\end{figure}

Figure \ref{fig:simple_anlgs} illustrates simple analogies that one could draw in any given RPM problem, where the images are represented by characters. These analogies are either between rows (Figure \ref{fig:simple_anlgs_h22} and \ref{fig:simple_anlgs_h33}) or between columns (Figure \ref{fig:simple_anlgs_v22} and \ref{fig:simple_anlgs_v33}), implying that the rows or columns share the same underlying relation among entries.

\begin{figure}[ht]
    \centering
    \begin{subfigure}{0.15\textwidth}
        \centering
        \includegraphics[width=\textwidth]{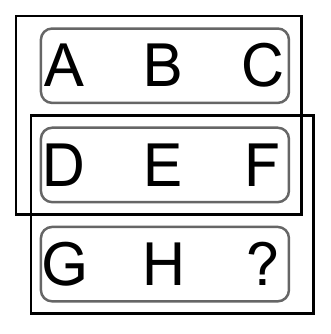} 
        \caption{}
        \label{fig:rcsv_anlg_a}
    \end{subfigure}
    \begin{subfigure}{0.15\textwidth}
        \centering
        \includegraphics[width=\textwidth]{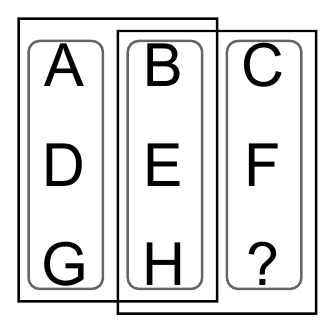} 
        \caption{}
        \label{fig:rcsv_anlg_b}
    \end{subfigure}
    \begin{subfigure}{0.15\textwidth}
        \centering
        \includegraphics[width=\textwidth]{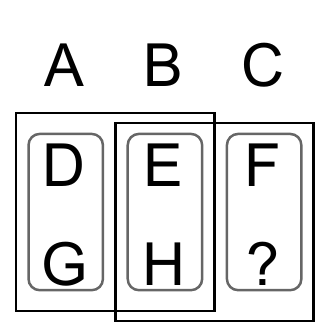} 
        \caption{}
        \label{fig:rcsv_anlg_c}
    \end{subfigure}
    \begin{subfigure}{0.15\textwidth}
        \centering
        \includegraphics[width=\textwidth]{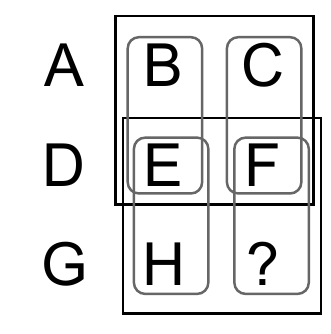} 
        \caption{}
        \label{fig:rcsv_anlg_d}
    \end{subfigure}
    \begin{subfigure}{0.15\textwidth}
        \centering
        \includegraphics[width=\textwidth]{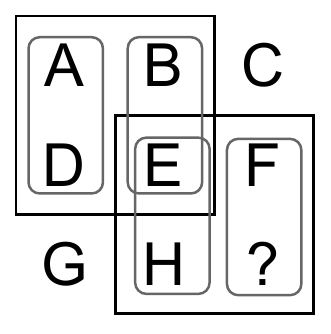} 
        \caption{}
        \label{fig:rcsv_anlg_e}
    \end{subfigure}
    
    \begin{subfigure}{0.15\textwidth}
        \centering
        \includegraphics[width=\textwidth]{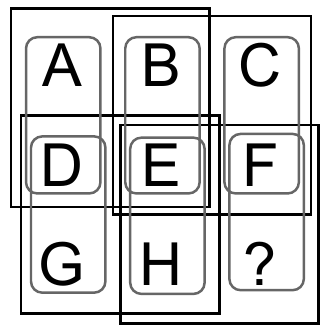} 
        \caption{}
        \label{fig:rcsv_anlg_f}
    \end{subfigure}
    \begin{subfigure}{0.15\textwidth}
        \centering
        \includegraphics[width=\textwidth]{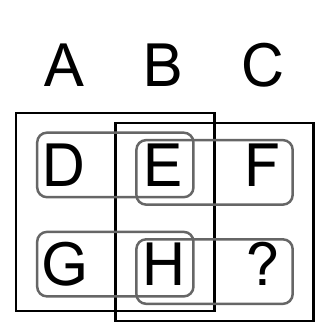} 
        \caption{}
        \label{fig:rcsv_anlg_g}
    \end{subfigure}
    \begin{subfigure}{0.15\textwidth}
        \centering
        \includegraphics[width=\textwidth]{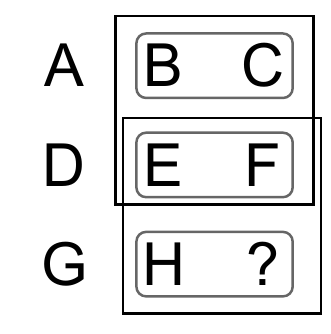} 
        \caption{}
        \label{fig:rcsv_anlg_h}
    \end{subfigure}
    \begin{subfigure}{0.15\textwidth}
        \centering
        \includegraphics[width=\textwidth]{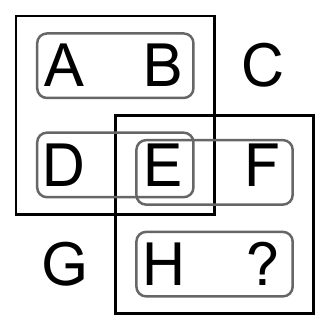} 
        \caption{}
        \label{fig:rcsv_anlg_i}
    \end{subfigure}
    \begin{subfigure}{0.15\textwidth}
        \centering
        \includegraphics[width=\textwidth]{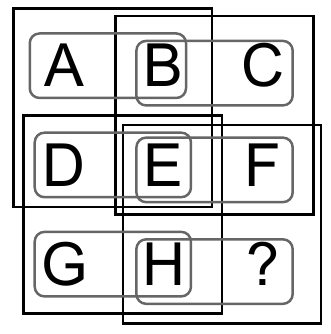} 
        \caption{}
        \label{fig:rcsv_anlg_j}
    \end{subfigure}
    \caption{Illustrations of recursive analogies in 3$\times$3 RPM problems: (a) are (b) are trio analogies and (c) through (j) are  pair analogies.}
    \label{fig:rcsv_anlg}
\end{figure}

In addition to the simple analogies in Figure \ref{fig:simple_anlgs}, the ASTI+ model also expands these analogies in two ways.  First, for $3\times3$ matrices, the model further considers several subproblems, as shown in Figure \ref{fig:rcsv_anlg}. For example, consider the simple analogies in Figure \ref{fig:simple_anlgs_h33}, \textit{A:B:C::G:H:?} and \textit{D:E:F::G:H:?}, which use only two of the three rows. We then combine them into a larger recursive\footnote{Recursive in that it is an analogy of analogies.} format, \textit{A:B:C::D:E:F:::D:E:F::G:H:?} as in Figure \ref{fig:rcsv_anlg_a}, which use all rows. In this recursive analogy, two subproblems are created --- the first subproblem is \textit{A:B:C::D:E:?}, with \textit{F} as the only option, and the second subproblem is \textit{D:E:F::G:H:?}, with options from the original RPM problem. All subproblems should be solved equally well by the correct transformation.

\begin{figure}[ht]
    \centering
    \begin{subfigure}{0.15\textwidth}
        \centering
        \includegraphics[width=\textwidth]{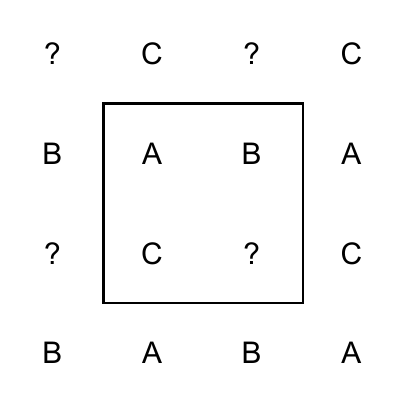} 
        \caption{}
        \label{fig:anlg_matrix_a}
    \end{subfigure}
    \begin{subfigure}{0.15\textwidth}
        \centering
        \includegraphics[width=\textwidth]{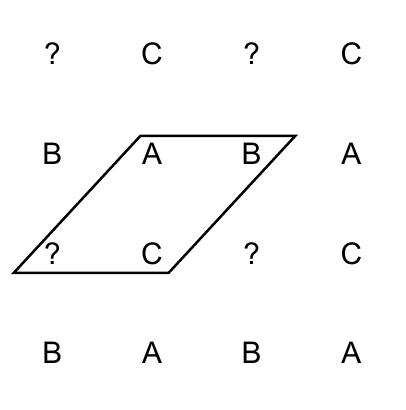} 
        \caption{}
        \label{fig:anlg_matrix_b}
    \end{subfigure}
    \begin{subfigure}{0.15\textwidth}
        \centering
        \includegraphics[width=\textwidth]{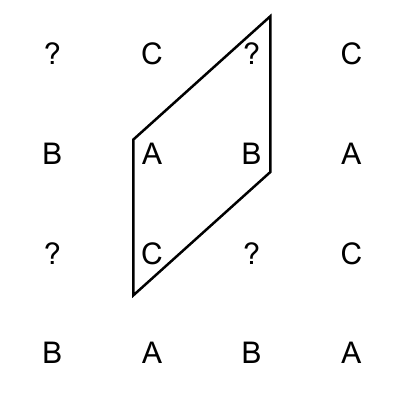} 
        \caption{}
        \label{fig:anlg_matrix_c}
    \end{subfigure}
    
    \begin{subfigure}{0.24\textwidth}
        \centering
        \includegraphics[width=\textwidth]{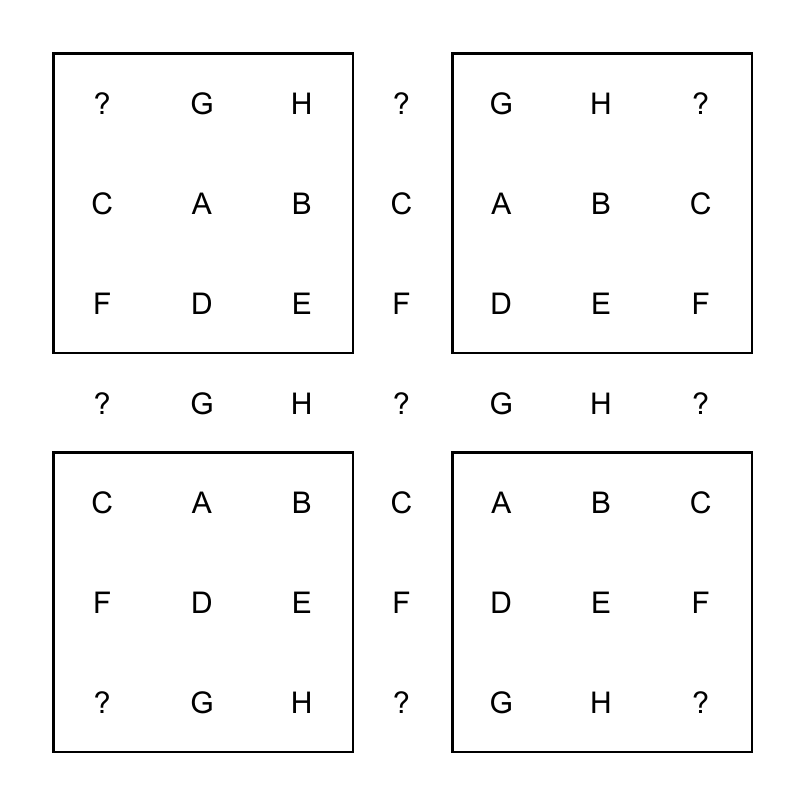} 
        \caption{}
        \label{fig:anlg_matrix_d}
    \end{subfigure}
    \begin{subfigure}{0.24\textwidth}
        \centering
        \includegraphics[width=\textwidth]{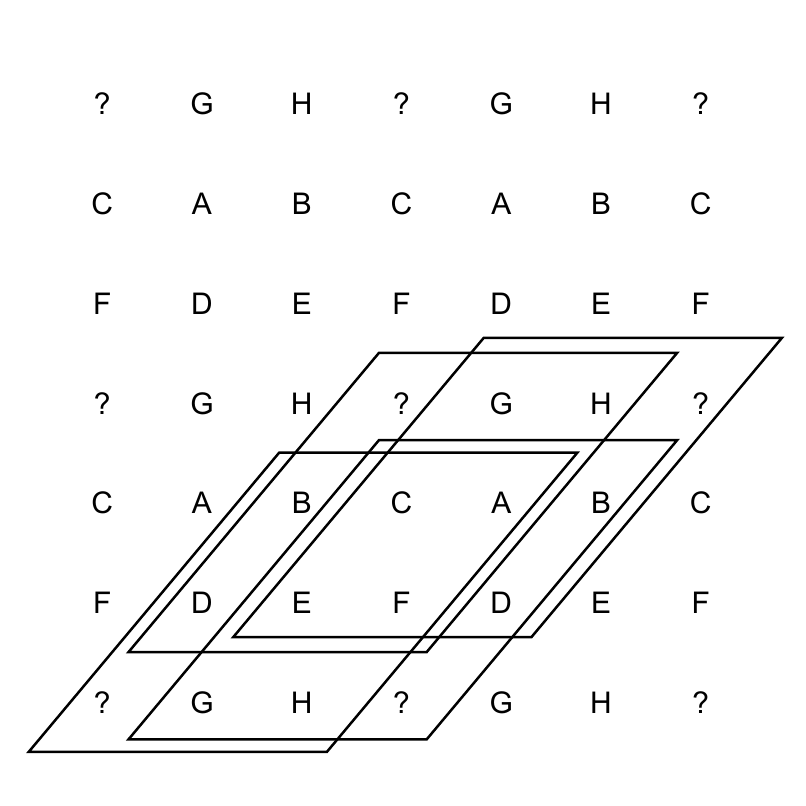} 
        \caption{}
        \label{fig:anlg_matrix_e}
    \end{subfigure}
    \begin{subfigure}{0.24\textwidth}
        \centering
        \includegraphics[width=\textwidth]{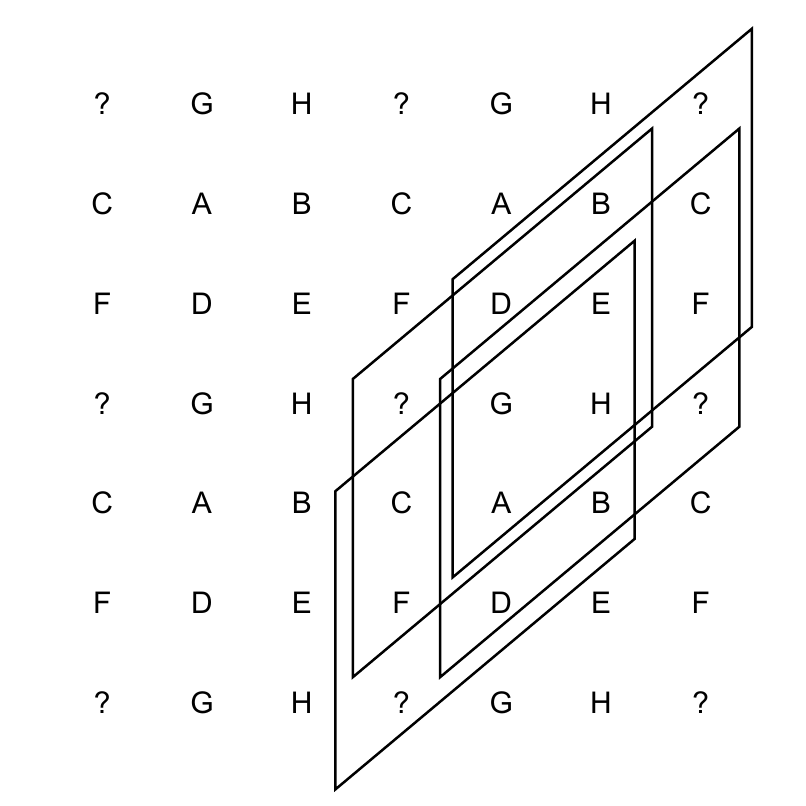} 
        \caption{}
        \label{fig:anlg_matrix_f}
    \end{subfigure}
    \begin{subfigure}{0.24\textwidth}
        \centering
        \includegraphics[width=\textwidth]{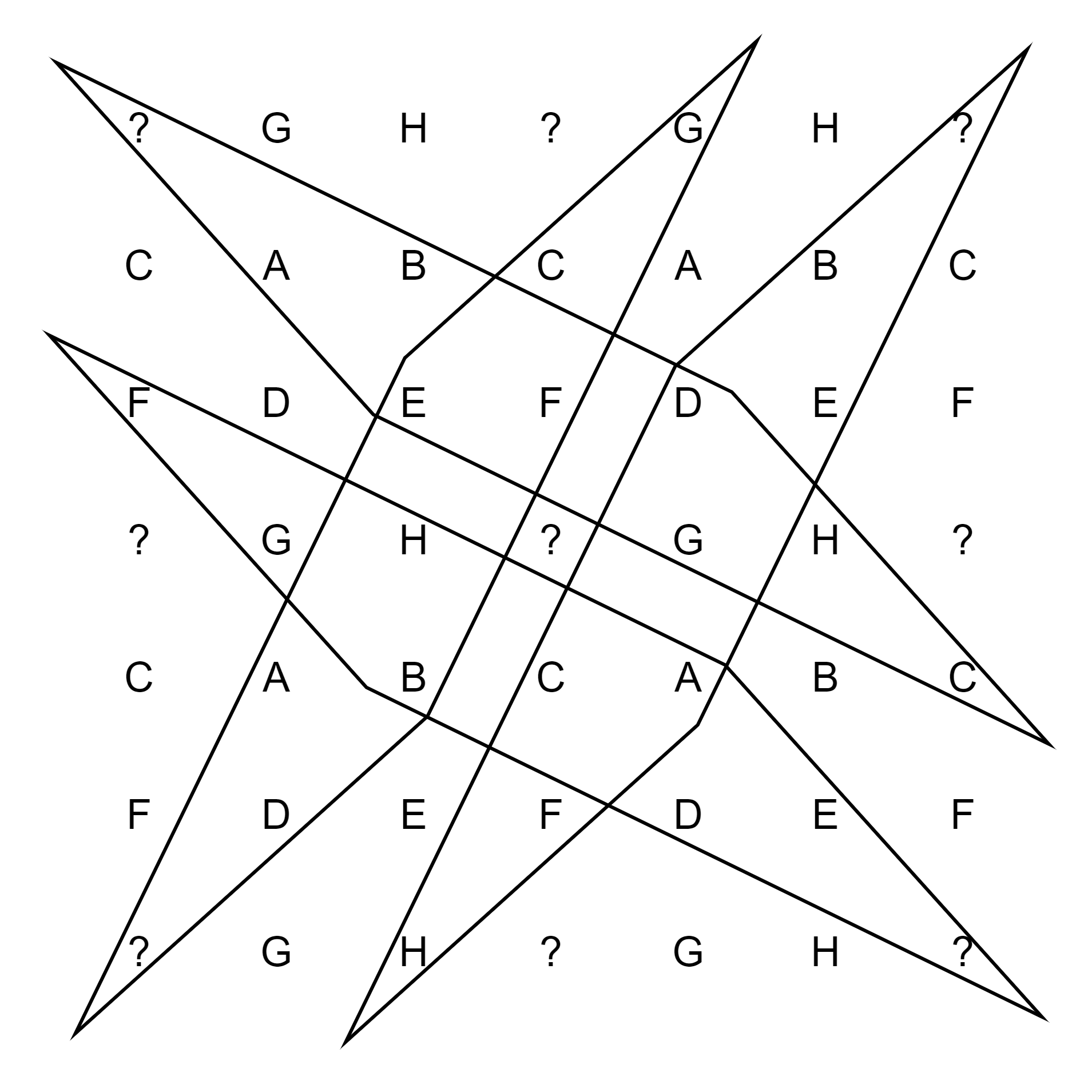} 
        \caption{}
        \label{fig:anlg_matrix_g}
    \end{subfigure}
    \caption{Expanded matrices to generate analogies: (a) through (c) are expanded from the 2$\times$2 matrix in Figure \ref{fig:simple_anlgs}, and (d) through (g) are expanded from the 3$\times$3 matrix in Figure \ref{fig:simple_anlgs}.}
    \label{fig:anlg_matrix}
\end{figure}

Second, ASTI+ captures more sophisticated spatial regularities by expanding the matrix in a way that the adjacency between matrix entries is preserved everywhere in the expanded version. Then it encloses different parts of the expanded matrix with quadrilaterals, as shown in Figure \ref{fig:anlg_matrix}. The entries in each quadrilateral form a new matrix, whose rows and columns constitute analogies that can not be systematically constructed by rows and columns in the original matrix. ASTI+ follows two reasonable heuristics to enclose these matrices: (1) the quadrilateral should contain a permutation of the original matrix, and (2) the quadrilateral should have a \textit{?} at one of its corners. We do not necessarily expect that humans use this strategy to search through this analogy space, but it provides a systematic and parsimonious way to capture regularities within a  matrix that humans might perceive and reason about, albeit in different ways.

\subsection{General Integration Strategy}
\label{sec:strgs}
The fourth element concerns the general strategy used to integrate transformations, analogies and similarity metrics to solve an RPM. The integration can be generally divided into three stages. In Stage 1, ASTI+ attempts to explain the variations in the incomplete matrix with some analogies and transformations. In Stage 2, it verifies the explanations by checking if there exists an option that can be generated from the analogy and transformation. In Stage 3, it uses the best explanation --- the best analogy and the best transformation --- to select an answer option.

To quantify ``how well'' an analogy and a transformation explain the variations across matrix entries, we introduce three scores corresponding to the three stages, which are realized through different ways to assemble Jaccard similarity measurements: (1) the $\mathsf{MAT}$ score measures how well an analogy and a transformation explain the variations in the matrix in Stage 1; (2) the $\mathsf{O}$ score measures how well an analogy and a transformation explain the variations involving the options in Stage 2; and (3) the $\mathsf{MATO}$ score, which is used as the final metric to select the answer, is computed from the $\mathsf{MAT}$  and $\mathsf{O}$ scores. For example, given the matrix in Figure \ref{fig:simple_anlgs_h22}, analogy \textit{A:B::C:?} and transformation $flip(X)$, we have $\mathsf{MAT}=J(flip(A),B)$, $\mathsf{O}=J(flip(C),O)$ and $\mathsf{MATO}=(\mathsf{MAT} + \mathsf{O}) / 2$. Score calculation depends on what types of analogy and transformation are used, as described below.

\textbf{\textsf{MAT} Scores.} 
For transformations in forms of $T(A)$ or $T(A,B)$ (without extra parameters), $\mathsf{MAT}$ scores are calculated in the same way as $flip(X)$. For transformations with extra parameters, they cannot be computed in this way because the model does not know the extra parameters. For example, for $add\_\mathit{diff}(I|S,T)$ and \textit{A:B::C:?}, it cannot use $\mathsf{MAT}=J(add\_\mathit{diff}(A|S,T), B)$ because it does not know $S$ and $T$, but it can use $add\_\mathit{diff}(C|A,B)$ to calculate $\mathsf{O}$ score. In this case, the $\mathsf{MAT}$ score is calculated as $\mathsf{MAT}=J_A(A,B)$ for $add\_\mathit{diff}(I|S,T)$. Although the model takes transformation-specific approaches to calculate $\mathsf{MAT}$ scores, they are simply different ways to assemble similarity measurements (symmetric and asymmetric Jaccard indices) of the same known matrix entries.

\textbf{\textsf{O} Scores.} 
For transformations whose $\mathsf{MAT}$ scores are calculated through the Jaccard index, so are their $\mathsf{O}$ scores. For transformations using the asymmetric Jaccard index, for example $add\_\mathit{diff}$ and $sub\_\mathit{diff}$, the asymmetric Jaccard index is always higher than the Jaccard index given the same input (see Equation \eqref{eq:jaccard} and \eqref{eq:asymmetric_jaccard}). As a result, transformations measured by asymmetric Jaccard index tend to have higher scores even if their explanations are poor. To fix this issue, the model calculates multiple Jaccard and asymmetric Jaccard indices, each of which characterizes a distinct aspect of the transformation,  and average them to get an $\mathsf{O}$ score. For example, for $add\_\mathit{diff}(C|A,B)$ and \textit{A:B::C:?}, three aspects of the transformation are considered: (1) how much $C$ is a subset of $O$, where $O$ is an option, (2) how the difference between $A$ and $B$ compares to the difference between $C$ and $O$ and (3) how similar the predicted image is to $O$. This leads to $\mathsf{O}=(J_A(C,O) + J(D,D') + J(add\_\mathit{diff}(C|A,B),O))) / 3$, where $D=B-A$ and $D'=O-C$ after $A$, $B$, $C$ and $O$ are properly aligned.

\textbf{\textsf{MATO} Scores.} Finally, every combination of an analogy, a transformation and an option is evaluated by a weighted average of its $\mathsf{MAT}$ and $\mathsf{O}$ scores, where the weight is proportional to the number of variations that the score measures. For recursive analogies in 3$\times$3 matrices, scores of the original problem are derived from the scores of subproblems. For instance, suppose that there are $n$ subproblems in a recursive analogy, and let $\mathsf{MAT}_k$ and $\mathsf{O}_k$ be the $\mathsf{MAT}$ score and $\mathsf{O}$ score of the $k$-th subproblem. In this case, the final $\mathsf{MAT}$ score is  $\mathsf{MAT} =[ \sum_{k=1}^{n-1} (\mathsf{MAT}_k + \mathsf{O}_k) + \mathsf{MAT}_n] / (2n - 1)$ and the final $\mathsf{MATO}$ score is $\mathsf{MATO}=[ \sum_{k=1}^{n} (\mathsf{MAT}_k + \mathsf{O}_k)] / 2n$.


 \begin{figure}[t]
    \centering
    \includegraphics[width=0.6\textwidth]{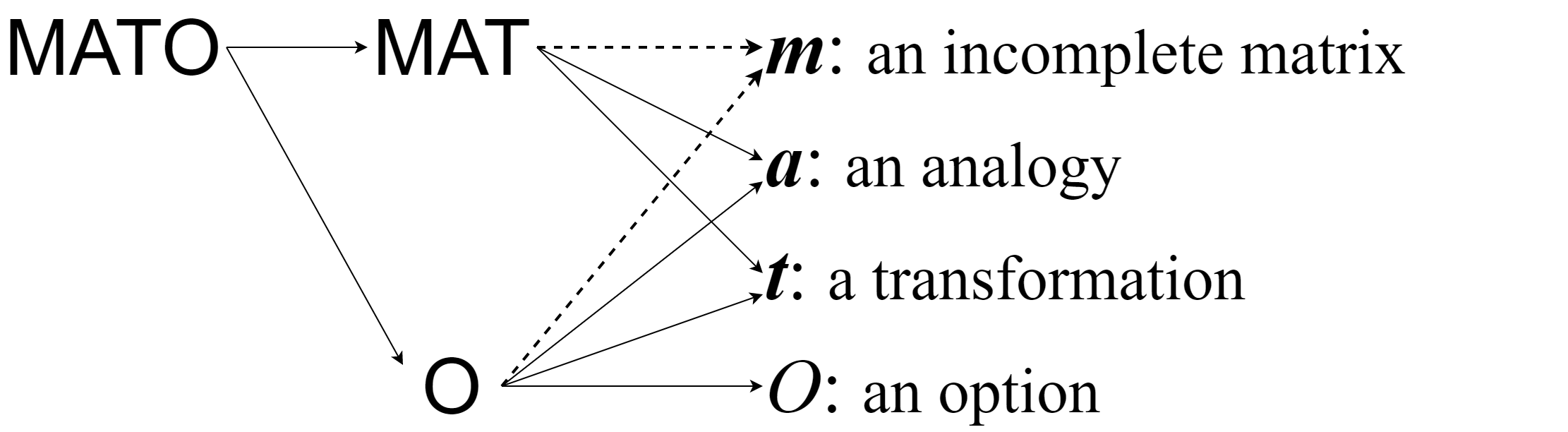}
    \caption{The dependencies of scores: The dashed lines denote partial dependence. Given the relations in an analogy, $\mathsf{MAT}$ relies on the entries that are not related to the missing entries while $\mathsf{O}$ relies on the entries that are related to the missing entries.}
    \label{fig:score_function_dependency}
\end{figure}

\subsection{Specific Integration Strategies: When and What to Maximize}
\label{sec:sis}

ASTI+ implements the general integration strategy as several alternative specific strategies that systematically explore different design choices in each stage of the general strategy. Given the dependencies of scores in Figure \ref{fig:score_function_dependency}, the general strategy boils down to an optimization in which $\mathsf{MATO}$ score is maximized over the analogy $\bm{a}$, the transformation $\bm{t}$, and the option $O$ for a problem-specific matrix $\bm{m}$. An heuristic for solving the optimization can be drawn from an observation on high-achieving human solvers --- they often first form a good understanding of the incomplete matrix before attending to the options. This observation, translated into our scoring system, says that a good $\mathsf{MAT}$ score implies a good $\mathsf{O}$ score and thus a good $\mathsf{MATO}$ score. 
However, as most heuristics in intelligent systems, this heuristic might become invalid in some cases, for example, it will not work if the system does not have adequate capability to fully ``understand'' or explain the incomplete matrix (e.g. lacking appropriate transformations or analogies), or if the matrix contains distracting noisy features that cause the system to ``over-explain'' the content that should have been ignored.

For this reason, we introduced specific integration strategies (summarized in the first part of Table \ref{table:configuration}) that range from relying entirely on the heuristic to ignoring it. In particular, given an RPM matrix $\bm{m}$, an analogy $\bm{a}$, a transformation $\bm{t}$ and an option $O$, the $\mathsf{MAT}$ score is a function $\mathsf{MAT}(\bm{m}, \bm{a}, \bm{t})$, $\mathsf{O}$ score is a function $\mathsf{O}(\bm{m}, \bm{a}, \bm{t},O)$, and $\mathsf{MATO}$ is a function $\mathsf{MATO}(\mathsf{MAT}, \mathsf{O})$. We formulate the three strategies as optimization processes, as shown below in \eqref{eq:cfdt}, \eqref{eq:ntrl} and \eqref{eq:prdt}:
\begin{align}
\label{eq:cfdt}
\begin{split}
     \mathsf{MATO}^* &= \max_{O} \mathsf{MATO}(\mathsf{MAT}(\bm{m}, \bm{a}^*, \bm{t}^*), \mathsf{O}(\bm{m}, \bm{a}^*, \bm{t}^*, O)) \\[-0.08in]
     \bm{a}^*, \bm{t}^* &= \argmax_{\bm{a}, \bm{t}} \mathsf{MAT}(\bm{m}, \bm{a}, \bm{t})
\end{split} 
\tag{$\mathsf{I}$}
\end{align}
\vskip -0.15in
\begin{align}
\label{eq:ntrl}
\begin{split}
     \mathsf{MATO}^* &= \max_{\bm{a}, O} \mathsf{MATO}(\mathsf{MAT}(\bm{m}, \bm{a}, \bm{t}^*), \mathsf{O}(\bm{m}, \bm{a}, \bm{t}^*, O)) \\[-0.08in]
     \bm{t}^* &= \argmax_{\bm{t}} \mathsf{MAT}(\bm{m}, \bm{a}, \bm{t}) 
\end{split} 
\tag{$\mathsf{II}$}
\end{align}
\vskip -0.15in
\begin{align}
\label{eq:prdt}
     \mathsf{MATO}^* &= \max_{\bm{a}, \bm{t}, O} \mathsf{MATO}(\mathsf{MAT}(\bm{m}, \bm{a}, \bm{t}), \mathsf{O}(\bm{m}, \bm{a}, \bm{t}, O))
\tag{$\mathsf{III}$}
\end{align} 
where \eqref{eq:cfdt} completely relies on the heuristic, \eqref{eq:prdt} completely ignores the heuristic, and \eqref{eq:ntrl} lies in between. We thus refer to optimizations \eqref{eq:cfdt}, \eqref{eq:ntrl} and \eqref{eq:prdt} as \textbf{M-confident}, \textbf{M-neutral} and \textbf{M-prudent} strategies, respectively, in the following discussion. 

Since the $\mathsf{O}$ score also depends on the option $O$ in Figure \ref{fig:score_function_dependency}, it can also serve as the objective function to select an answer from the options. Therefore, ASTI+ has three analogous integration strategies for maximizing $\mathsf{O}$, which we refer to as \textbf{O-confident}, \textbf{O-neutral} \eqref{eq:ntrl_O} and \textbf{O-prudent} \eqref{eq:prdt_O}) strategies:

\begin{align}
\label{eq:ntrl_O}
\begin{split}
     \mathsf{O}^*&=\max_{\bm{a}, O} \mathsf{O}(\bm{m}, \bm{a}, \bm{t}^*, O) \\[-0.08in]
      \bm{t}^*&=\argmax_{\bm{t}} \mathsf{MAT}(\bm{m}, \bm{a}, \bm{t})
\end{split} 
\tag{$\mathsf{IV}$}
\end{align}
\vskip -0.2in
\begin{align}
\label{eq:prdt_O}
     \mathsf{O}^*&=\max_{\bm{a}, \bm{t}, O} \mathsf{O}(\bm{m}, \bm{a}, \bm{t}, O)
\tag{$\mathsf{V}$}
\end{align} 
Note that $\mathsf{MATO}$ is simply a weighted average of $\mathsf{MAT}$ and $\mathsf{O}$, so the O-confident strategy is equivalent to M-confident \eqref{eq:cfdt}. Thus we do not need a separate optimization for it.

\subsection{From ASTI to ASTI+}
In this subsection, we compare ASTI+ to its predecessor ASTI. The ASTI model \citep{kunda2013computational,kunda2013visual} introduced a visual-imagery framework for solving geometric reasoning problem that based analogical reasoning on a pixel-level representation, transformations, and metrics. This framework remains unchanged in the ASTI+ model. From ASTI to ASTI+, we gave enhancements to the core dimensions of the framework.

\paragraph{Analogy} For 2$\times$2 matrices, ASTI and ASTI+ share the same analogy set, which could be manually enumerated given the small size of matrices. In contrast, 3$\times$3 matrices provide many more choices of analogies. We thus developed the systematic approach in Section \ref{sec:anlgs} to enumerate analogies, which led to analogies that ASTI supported. We adopted this approach because analogues  in matrix reasoning tasks are usually arranged in spatial parallelism. Another enhancement was the introduction of recursive analogy, which was inspired by the recursive and incremental nature of human solving reported in the literature \citep{carpenter1990one, kunda2015computational}.

\paragraph{Transformation} ASTI+ inherits all the affine transformations of ASTI. Meanwhile, ASTI+ has extra complex set operations, such as \textit{inverse unite} and \textit{shadow mask unite}, that combine basic set operations in ASTI. 

\paragraph{Integration Strategy} Compared to ASTI, ASTI+ has more choices of integration strategy representing different degrees of reliance on the heuristic mentioned in Section \ref{sec:sis}. In contrast, ASTI implements only one strategy that roughly equals the \textbf{M-prudent} strategy in ASTI+. 

\paragraph{Option-Usage Strategy} Two general option-usage strategies for solving RPM problems and other multiple-choice reasoning problems have been reported in human studies: constructive matching and response elimination \citep{snow1980aptitude, bethell1984adaptive}. Constructive matching proceeds as $infer(T) \rightarrow A \myeq apply(T) \rightarrow test(A)$, where $T$ is a transformation and $A$ is an answer constructed by applying $T$. Response elimination proceeds as $infer(T_1) \rightarrow infer(T_2 | O) \rightarrow compare(T_1, T_2)$, where $O$ is an option used to infer $T_2$ and, if $compare(T_1, T_2)$ fails, $O$ will be eliminated. The strategy choice observed in human experiments was found to relate to subject's intellectual ability, item type and difficulty. Cognitive models have been constructed based on both strategies \citep{evans1964program, sternberg1977intelligence, mulholland1980components}.

ASTI strictly follows the constructive matching strategy, where options are never used before generating the missing entry. We refer to this constructive matching as \textbf{option-free constructive matching}. In contrast, ASTI+ adopts a slightly different approach that we refer to as \textbf{option-informed constructive matching}, which lies between constructive matching and response elimination. It follows the pattern $infer(T, p_1) \rightarrow infer(p_2 | T, O)\: \& \: A \myeq apply(T, p_2) \rightarrow evaluate(A, O, p_1, p_2)$, where $p_1$ and $p_2$ are parameters of $T$ and $p_2$ is inferred from the option $O$. For example, options are used for calculating alignment parameters of the transformations in ASTI+. This strategy gives ASTI+ the flexibility to represent the relations that cannot be represented by single-direction transformations.

\section{Experimental Studies of the ASTI+ Model}
\label{sec:exprs}

To study how different analogical constructions affect the performance on the RPM test, we equip ASTI+ with different configurations of analogies and transformations, and integration strategies, and test its performance on the standard RPM test, which consists of five sets of problems with 12 problems each. In our experiments, the analogical constructions are implemented as different configurations of analogies, transformations, and integration strategies. We further aggregated them into the groups summarized in Table \ref{table:configuration}. Each configuration has one or more groups of analogies and transformations, whereas it has only one integration strategy. We hypothesized that, by varying the configuration, the performance would change accordingly. 

To study how each dimension of the configuration affects performance, we conducted two experiments. In the first one, we varied only the integration strategy and fixed the configuration of analogies and transformations (using the full set of analogies and transformations). In the second one, we selected the best integration strategy in the first one and varied the configurations of analogies and configurations.

\begin{table}[t]
    \centering
    \caption{Configurations of integration strategies, analogy groups and transformation groups.}
    \label{table:configuration}
\begin{threeparttable}
    \textsf{
        \rule{\textwidth}{0.4pt}
        \scriptsize{
            \begin{compactlabel}{\textbf{M-confident}}
                \item[\textbf{Integration Strategies}]
                \item[{M-Confident}] Find an analogy and a transformation that best explain the incomplete matrix; and then select an option that best matches the analogy and the transformation.
                \item[{O-Confident}] Mathematically equivalent to \textsf{M-confident}.
                \item[{M-neutral}] For each analogy, find a transformation that best explains the incomplete matrix; and then select an option such that there exist an analogy and its best transformation that match the option well and explain the incomplete matrix well.
                \item[{O-neutral}] For each analogy, find a transformation that best explains the incomplete matrix; and then select an option such that there exist an analogy and its best transformation that match the option well.
                \item[{M-prudent}] Select an option such that there exist an analogy and a transformation that match the option well and explain the incomplete matrix well.
                \item[{O-prudent}] Select an option such that there exist an analogy and a transformation that match the option well.
            \end{compactlabel}
        }
        \rule{\textwidth}{0.4pt}
        \scriptsize{
            \begin{compactlabel}{\textbf{M-confident}}
                \item[\textbf{Transformation Groups}]
                \item[{Affine}] All the affine transformations.
                \item[{Diff}] $add\_\mathit{diff}$, $sub\_\mathit{diff}$, $xor\_\mathit{diff}$, and $preserving\_sub\_\mathit{diff}$.
                \item[{Match}] $duplicate$ and $rearrange$.
                \item[{Set}] $unite$, $intersect$, $inverse\_unite$, $xor$ and $shadow\_mask\_unite$.
            \end{compactlabel}
        }
        \rule{\textwidth}{0.4pt}
        \scriptsize{
            \begin{compactlabel}{\textbf{M-confident}}
                \item[\textbf{Analogy Groups}]
                \item[{S}] The analogies in Figure \ref{fig:anlg_matrix_a} and \ref{fig:anlg_matrix_d}.
                \item[{H}] The analogies in Figure \ref{fig:anlg_matrix_b} and \ref{fig:anlg_matrix_e}.
                \item[{V}] The analogies in Figure \ref{fig:anlg_matrix_c} and \ref{fig:anlg_matrix_f}.
                \item[{R}] The analogies in Figure \ref{fig:anlg_matrix_g}. 
            \end{compactlabel}
        }
        \rule{\textwidth}{0.4pt}
    }
\end{threeparttable}
\end{table}

\begin{figure}[t]
    \centering
    \begin{subfigure}{0.49\textwidth}
        \centering
        \includegraphics[width=\textwidth]{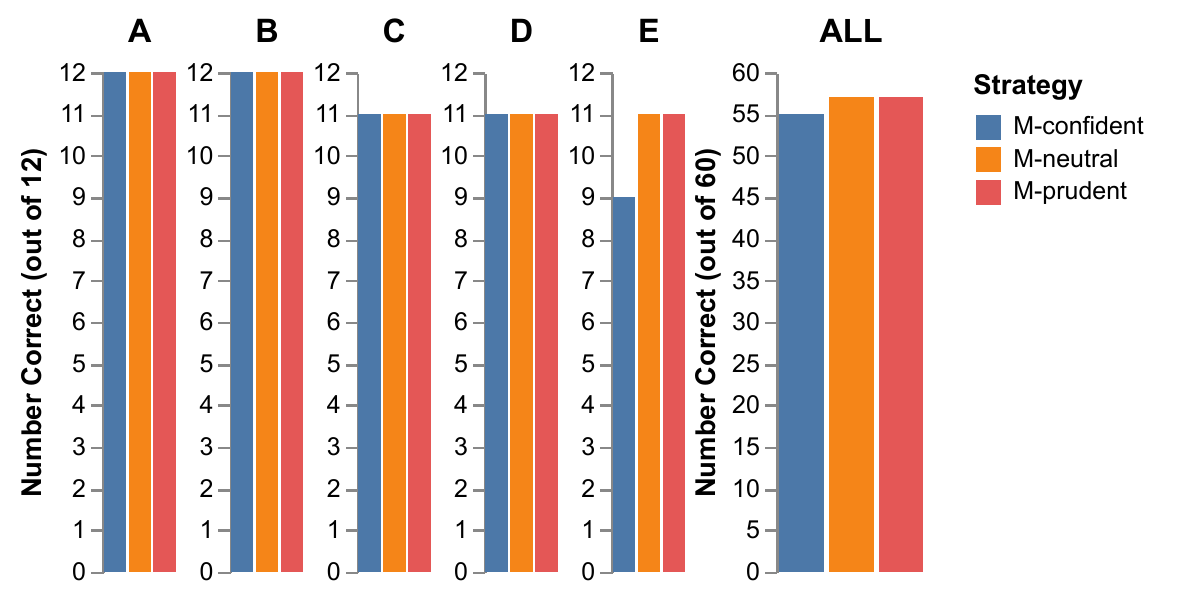} 
        \vskip -0.1in
        \caption{}
        \label{fig:strg_comparison_a}
    \end{subfigure}
    \begin{subfigure}{0.49\textwidth}
        \centering
        \includegraphics[width=\textwidth]{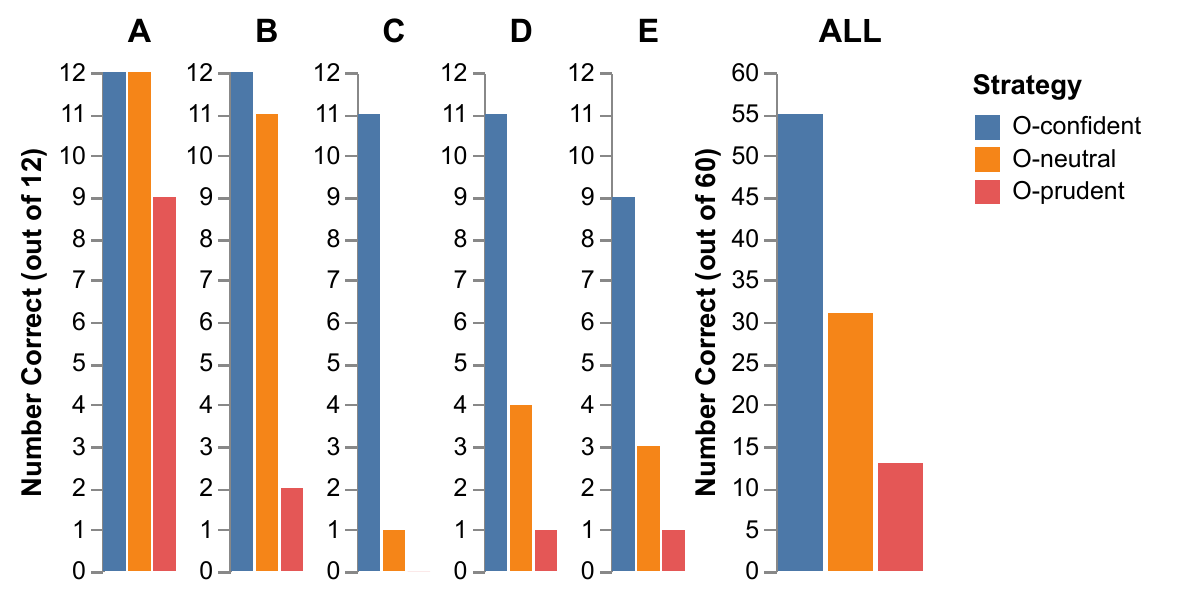} 
        \vskip -0.1in
        \caption{}
        \label{fig:strg_comparison_b}
    \end{subfigure}
    
    \begin{subfigure}{\textwidth}
        \centering
        \includegraphics[width=\textwidth]{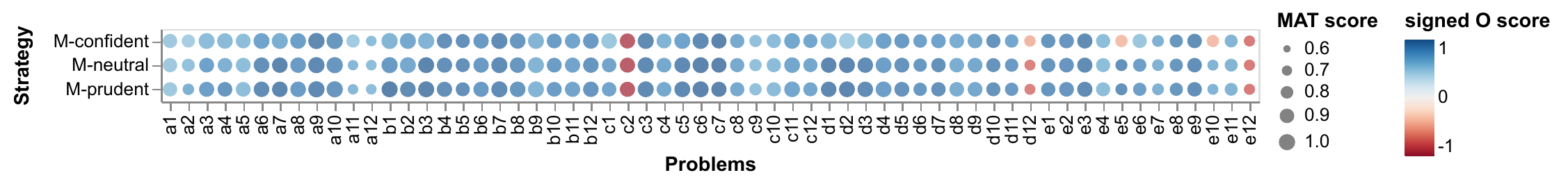} 
        \vskip -0.1in
        \caption{}
        \label{fig:strg_comparison_c}
    \end{subfigure}
    \begin{subfigure}{\textwidth}
        \centering
        \includegraphics[width=\textwidth]{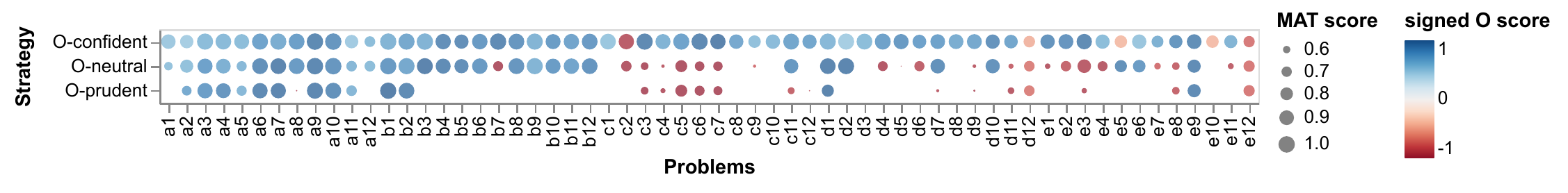} 
        \vskip -0.1in
        \caption{}
        \label{fig:strg_comparison_d}
    \end{subfigure}
    \vskip -0.15in
    \caption{Performance of each strategy on the standard RPM test: (a) and (b) show numbers of problems correctly solved by each strategy in every set (A---E) and the entire test; (c) and (d) visualize $\mathsf{MAT}$ and $\mathsf{O}$ scores of each strategy's answer to each problem as disks of various sizes and colors, where red disks indicates incorrect answers and blue disks indicate correct answers. Note that the ``signed'' $\mathsf{O}$ score in (c) and (d) is only to distinguish visually between correct and incorrect answers, and the real scores always fall in $[0, 1]$.}
    \label{fig:strg_comparison}
\end{figure}

\begin{figure}[t]
    \centering
    \begin{subfigure}{0.49\textwidth}
        \centering
        \includegraphics[width=\textwidth]{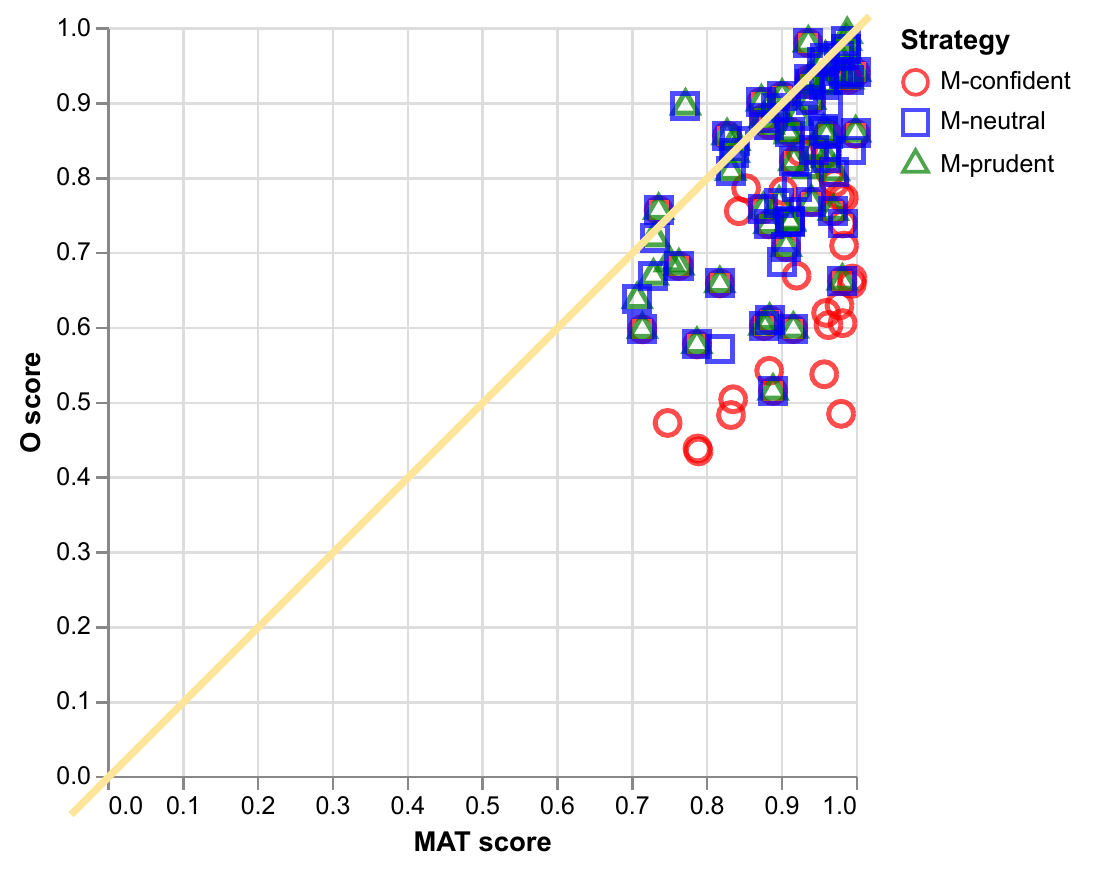} 
        \vskip -0.1in
        \caption{}
        \label{fig:strg_comparison_e}
    \end{subfigure}
    \begin{subfigure}{0.49\textwidth}
        \centering
        \includegraphics[width=\textwidth]{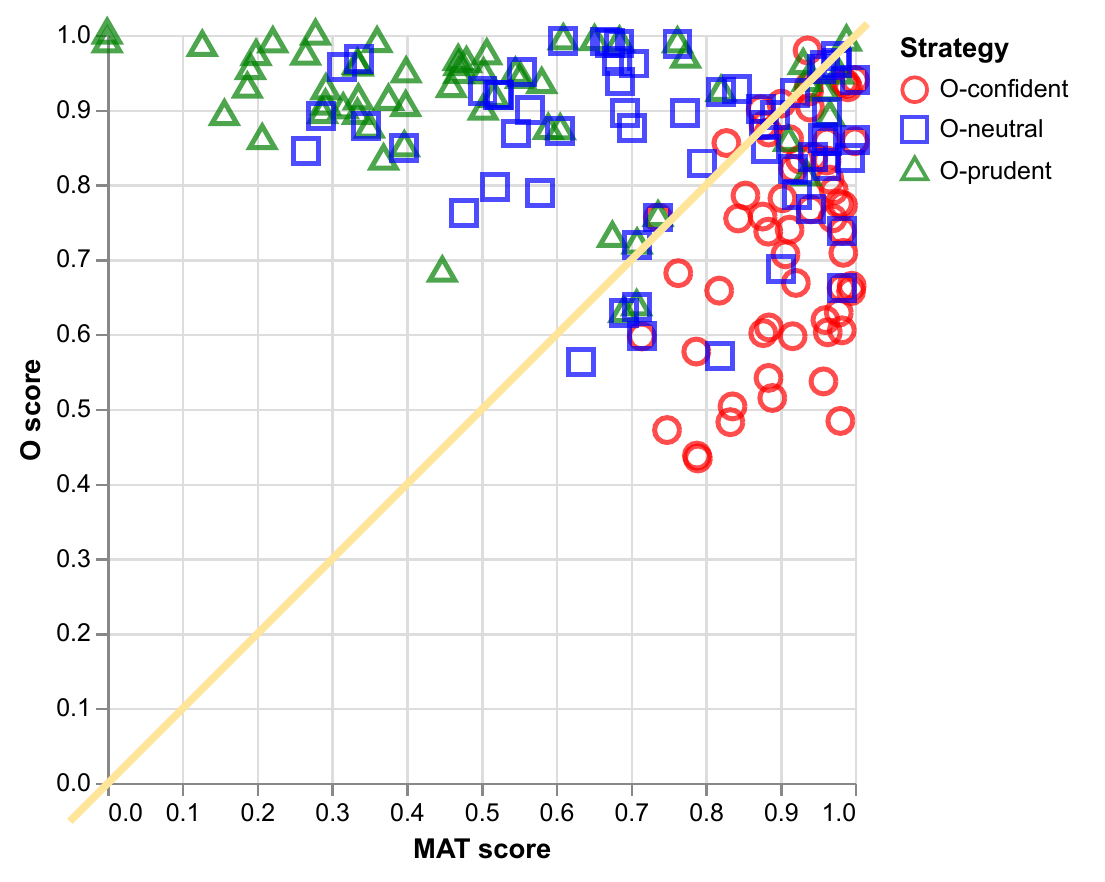} 
        \vskip -0.1in
        \caption{}
        \label{fig:strg_comparison_f}
    \end{subfigure}
    \vskip -0.15in
    \caption{Scatter plots of each strategy's answer to each problem in the standard RPM test drawn with respect to the $\mathsf{MAT}$ and $\mathsf{O}$ scores.}
    \vskip -0.15in
    \label{fig:strg_comparison_ef}
\end{figure}

The first experiment compares the integration strategies. Figure \ref{fig:strg_comparison} shows the set-wise and problem-wise performance of each one. The M-neutral strategy always ties with the M-prudent strategy, solving 57/60 problems, whereas the M-confident strategy performs slightly worse, solving 55/60 problems. The O strategies are far less capable, especially in the last three sets (C, D and E), where the problems are 3$\times$3 (Set A and B contains only 2$\times$2 problems). Thus, the M strategies, by considering both $\mathsf{MAT}$ and $\mathsf{O}$ scores, are more robust to increases in the matrix dimension.

While the M-confident comes in last in Figure \ref{fig:strg_comparison_a} by maximizing $\mathsf{MATO}$, the O-confident fares best by maximizing $\mathsf{O}$ in Figure \ref{fig:strg_comparison_b}. Furthermore, $\mathsf{O}$-neutral and $\mathsf{O}$-prudent strategies in Figure \ref{fig:strg_comparison_b} contrast sharply with their counterparts in Figure \ref{fig:strg_comparison_a}. In particular, the more a strategy relies on the heuristic from Section \ref{sec:sis}, the more performance drops when switching from maximizing $\mathsf{MATO}$ to maximizing $\mathsf{O}$.  We surmise that this is because the RPM is designed to have distractors with high $\mathsf{O}$ and low $\mathsf{MAT}$. In other words, these distractors work like traps for strategies that maximize only $\mathsf{O}$ scores, which is consistent with observations that people often make errors of ``repetition'' while solving RPM problems \citep{kunda2016error}.

Figure \ref{fig:strg_comparison_c} and \ref{fig:strg_comparison_d} depict the scores for each strategy's answer to each problem as disks. $\mathsf{MAT}$ and $\mathsf{O}$ scores are encoded as size and color intensity, while the correctness of the answer is denoted by colors (blue for correct and red for incorrect). Note that the ``signed'' $\mathsf{O}$ score in Figure \ref{fig:strg_comparison_c} and \ref{fig:strg_comparison_d} is only to distinguish between correct and incorrect answers, and the real scores always fall in $[0, 1]$. Figure \ref{fig:strg_comparison_c} and \ref{fig:strg_comparison_d} show a subtle difference: different strategies can have the same correct answer to a problem, but the answer may result from different analogies and transformations. Otherwise blue disks in any column would have the same size and color.

Figure \ref{fig:strg_comparison_e} and \ref{fig:strg_comparison_f} present the strategies' answers to every problem in scatter plots drawn with respect to the $\mathsf{MAT}$ and $\mathsf{O}$ scores, which show more difference between strategies. Note that most data points in Figure \ref{fig:strg_comparison_e}, corresponding to the blue disks in Figure \ref{fig:strg_comparison_c}, denote problems that are correctly solved. Since these data points in Figure \ref{fig:strg_comparison_e} are mostly located near or below the diagonal, we could hypothesize that, for a "naive" participant or computational model (with little prior knowledge about RPM), a good explanation for the known matrix entries matters more than how an option can be matched.  Recall that $\mathsf{MAT}$ and $\mathsf{O}$ are measurements of these two explanations. On the flip side, many more points, representing incorrect answers according to Figure \ref{fig:strg_comparison_d}, fall above the diagonal in Figure \ref{fig:strg_comparison_f}, which further supports this idea. The hypothesis is consistent with observations in previous human studies that high-achieving test takers usually take a more constructive approach, which requires a clear explanation of the matrix rather than perceptually matching the options \citep{bethell1984adaptive,carpenter1990one, lovett2017modeling}.

\begin{figure}[t]
    \centering
    \begin{subfigure}{\textwidth}
        \centering
        \includegraphics[width=\textwidth]{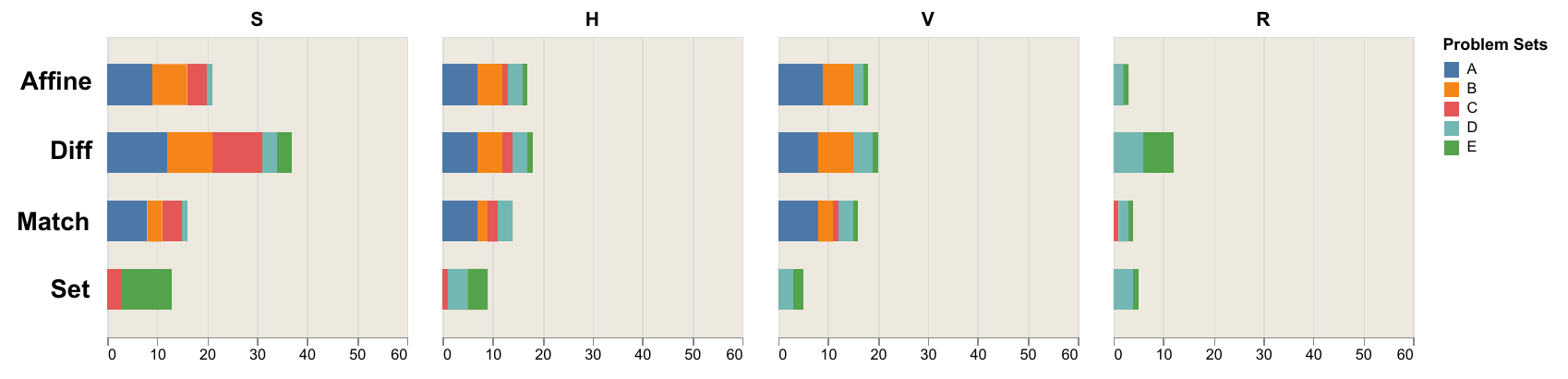} 
        \caption{}
        \label{fig:stack_bar_a}
    \end{subfigure}
    \begin{subfigure}{\textwidth}
        \centering
        \includegraphics[width=\textwidth]{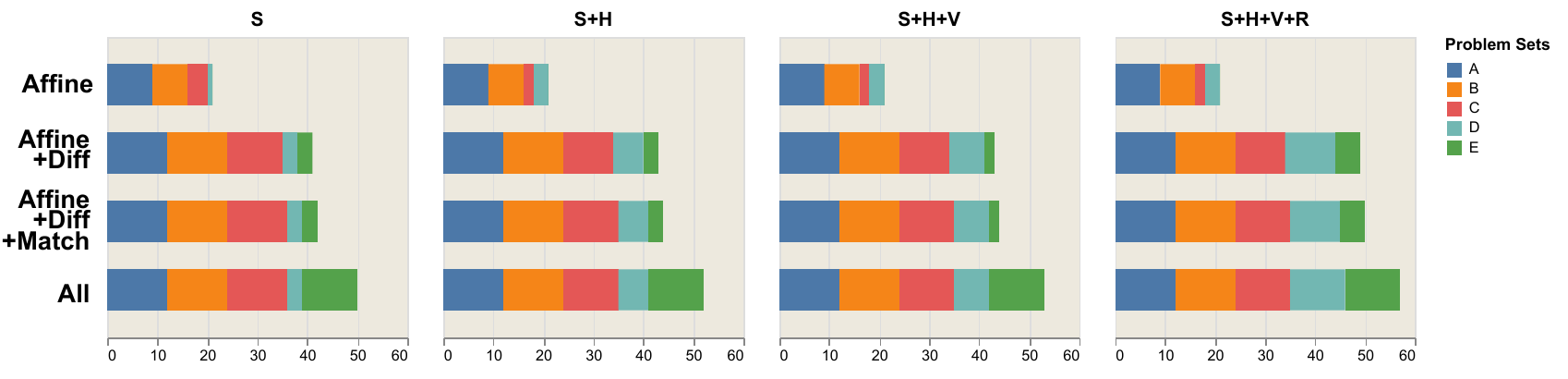} 
        \caption{}
        \label{fig:stack_bar_b}
    \end{subfigure}
    \caption{Bar charts of numbers of problems correctly solved by $\mathsf{M}$-prudent strategy using different analogy groups and transformation groups. (This figure should be viewed in color.)}
    \label{fig:stack_bar}
\end{figure}

In the second experiment, we compared different configurations of analogies and transformations while setting the integration strategy to the M-prudent strategy. Figure \ref{fig:stack_bar} shows the performance of different combinations of analogy and transformation groups in this situation. In particular, each analogy group is combined with each transformation group in  Figure \ref{fig:stack_bar_a}, and analogy groups and transformation groups are combined in an incremental way in Figure \ref{fig:stack_bar_b}. In Figure \ref{fig:stack_bar_a}, we can see the strength and weakness of each analogy group and each transformation group. \textsf{S} analogies plus \textsf{Diff} transformations are good at problems in Set A, B, and C, whereas \textsf{R} analogies and \textsf{Set} transformations work well on Set D and E but work poorly on Set A, B, and C. 
Figure \ref{fig:stack_bar_b} shows increases in both the vertical and horizontal directions. The former are more substantial than the latter. This does not mean that transformations are more important than analogies, because, as seen in Figure \ref{fig:stack_bar_a}, the \textsf{S} group outperforms \textsf{H}, \textsf{V}, and \textsf{R} for every transformation group and most problems in Set A, B and C solved by \textsf{H}, \textsf{V}, and \textsf{R} can also be also solved by \textsf{S}  with a different transformation. We might expect more variation across analogy groups if they were defined at a finer-grained level. 

\begin{figure}[htbp]
    \centering
    \includegraphics[width=\textwidth]{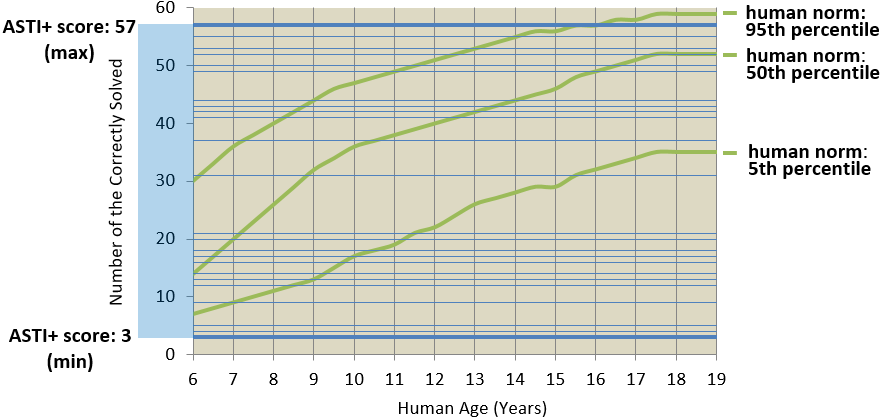}
    \caption{Comparison between ASTI+ and human subjects. The blue horizontal lines denote the performance of the configurations used in our experiments. The green curves represent the percentiles of human data.}
    \label{fig:human_cpr_2}
\end{figure}

To conclude our analysis, we compare ASTI+'s performance with human performance \citep{raven1998manual} (i.e., the normative data of RPM). Figure \ref{fig:human_cpr_2} shows 95th, 50th and 5th percentiles of human performance (age from 6 to 19) in green curves and the performance of different configurations of ASTI+ used in our experiments in horizontal blue lines\footnote{Note that we did not use all the possible configurations, which would have resulted in wider and more even distribution of blue lines}. The ranges of ASTI+'s and human performance overlap substantially, suggesting that analogical construction has a great effect on the performance of the RPM test and that it probably has the same effect on the performance in other geometric reasoning tasks.

\section{Related Work}

In this section, we examine the related research about reasoning on geometric matrix tasks. We first review the commonly used problem sets in this area, then discuss problem representations adopted by different computational models. We further analyze how analogies among matrix entries are interpreted differently in different problem sets and computational models. Finally, we compare the protocols used to evaluate the computational models and the integration strategies.

\paragraph{Problem Sets} 
The commonly used problem sets in geometric matrix reasoning task can be classified into three categories according to their original purpose. The first contains the problem sets that are handcrafted by psychologists and psychometricians and then applied to measure human intellectual abilities. The best-known examples are the three versions of the RPM test (colored, standard, and advanced) \citep{raven1998manual} and the Thurstone's problem set (a.k.a. Evans' problem set \citep{evans1964program}), which was published in the 1942 edition of the Psychological Test for College Freshman of the American Council on Education. The second category includes the problem sets that are automatically generated by computer programs based on the designs in the first category and are mostly used in large-scale and adaptive testing of human intellectual abilities \citep{yang2021automatic}. The third category includes the problem sets that are also automatically generated but used to train and evaluate machine learning models for abstract reasoning. The problem sets in this category are usually large-scale data sets of homogeneous matrix problems, whose psychometric qualities are usually not guaranteed as in the first two categories. The representatives of this category are Procedurally Generated Matrices (PGM) \citep{santoro2018measuring} and Relational and Analogical Visual rEasoNing (RAVEN) \citep{zhang2019raven}.

\paragraph{Problem Representations}
Problem representations have been generally grouped into two categories --- propositional and visual representations --- in previous studies
\citep{nersessian2010creating, kunda2013computational}. However, given the complex structure of RPM, different aspects of RPM can be represented differently, and thus it is inaccurate to say that an model is using a propositional or visual representation. Particularly, geometric objects, relations among geometric objects, and analogies between groups of geometric objects are represented differently in previous computational models. For example, the early models \citep{evans1964program, hunt1974quote, carpenter1990one, lovett2017modeling} used propositional representations for both the geometric objects and relations, and implicitly represented analogies by specific procedural arrangement of computation according to specific analogy interpretation, which will be discussed later. The machine learning models \citep{santoro2018measuring, zhang2019raven} visually represent the geometric objects, implicitly represent the relations through learned parameters, and implicitly represent the analogies through specific structures of machine learning models according to their assumed analogy interpretations. Our model and its predecessor \citep{kunda2013computational,kunda2013visual} take a different approach by representing goemetric objects visually and relations among them propositionally, and we encode analogies explicitly in a proportional format. Another aspect that one could consider is whether geometric objects are represented hierarchically \citep{cirillo2010anthropomorphic}. Rather than representing each matrix entry separately, one could also represent the entire RPM matrix as a single image \citep{hua2019modeling}, thus encoding every aspect visually and implicitly.

\paragraph{Analogy Interpretation}
Although an analogy is usually verbalized as ``\textit{A} is to \textit{B} as \textit{C} is to \textit{D}'' (i.e., \textit{A:B::C:D}), this must be interpreted quite differently in different tasks. As a result, computational models require specific assumptions about analogy interpretation. For example, RPM matrices have independent vertical and horizontal goemetric variations, and thus can be solved using either row or column analogies. In contrast, the Thurstone's problems have goemetric variations in only one direction, say row direction, and thus can only be solved using row analogies. In this case, the analogical relations between columns are naturally determined when a meaningful row analogy is determined for the problem. PGM and RAVEN problems are the same as Thurstone set in this regard. This closely relates to the central permutation property of analogy, i.e., \textit{A:B::C:D} if and only if \textit{A:C::B:D}. This property is true when the analogy has already been established in a specific context, for example, \textit{engine:fuel::human:food} versus \textit{engine:human::fuel:food}. But when a human or computational model is tasked to make an analogy with these four words, the thinking/solving process and difficulty can be quite different for these two analogies. Besides analogical directions, \textit{A:B::C:D} is also interpreted differently in other dimensions, such as ``\textit{A} is similar to \textit{B} as \textit{C} is similar to \textit{D}'' versus ``\textit{A} is different from \textit{B} as \textit{C} is different to \textit{D}'' \citep{prade2014homogenous}.


\paragraph{Transformations and Learning}
Transformations of geometric elements in computational cognitive models for solving RPM are usually predefined based on the observations on existing handcrafted problems \citep{hornke1986rule, carpenter1990one}. These transformations are very limited and usually do not go beyond progressions, arithmetic operation, and set/logical operation, which are defined on specific geometric elements. In addition, diversity and complexity of perception organization \citep{primi2001complexity} of these transformations and geometric elements were kept at a low level in current problem sets and computational models. For example, when multiple transformations are present in a problem, they are often manifested by independent or separate geometric elements, but the situation where one transformation depends on another is very rare. In machine learning models for solving RPM, transformations are learned from data sets while the models are trained to solve RPM. It has been shown that machine learning models are capable of learning the above transformations when they are bound with specific geometric elements \citep{santoro2018measuring, zhang2019raven}. But it is still questionable whether they learn the abstract concepts of transformations and analogy per se, which are conceptually invariant to geometric elements, because no good generalization across geometric elements has been reported.

\paragraph{Evaluation Protocols and Integration Strategies}
\citeauthor{benny2021scale} identified two typical evaluation protocols for computational models for solving RPM --- multiple choice and single choice. In single choice, the model scores each option independently and the option with the highest score is taken; in multiple choice, the model is allowed to score all options by comparing them. The integration strategies of ASTI+ could be considered as extensions of single choice evaluation.  On one hand, the option variable is always in the outermost layer of optimization in ASTI+ (as in single choice, select the highest-scored option), on the other, the integration strategies have more variables to arrange --- analogy and transformation variables --- in optimization (thus extending the protocol by providing extra choices of individually scoring each analogy and each transformation or comparing them, i.e., extra layers of single/multiple choice).

\section{Concluding Remarks}

This paper described a framework of solving geometric matrix reasoning tasks, including variations in transformations, analogies, and integration strategies.  We showed that this task-specific language of representations and inference mechanisms is quite expressive on the Raven's Standard Progressive Matrices test. We further demonstrated that test performance varies not only as a function of transformations and analogies used, but also with the higher-level integration strategy: when and how, across analogies and transformations, the model performs its maximization calculations. 

In tasks such as the RPM, where \textit{eductive ability} \citep{spearman1923nature,raven1998manual} is required to extract information from a new situation, redundant information often exists; otherwise, ambiguity cannot be eliminated because little prior knowledge is available. Methods for representing, identifying, and exploiting such redundancies are crucial to solving the problem. Analogy is often used for this purpose. By varying the configuration of the ASTI+ model, we alter its ability to identify and represent these redundancies and control the extent to which it can exploit them to solve the task. 

Our work has two main implications. First, for artificial intelligence, analogical ability might be needed for systems in new unseen situations.  Second, for human intelligence, understanding analogical ability helps us understand eductive ability. ASTI+ demonstrates that analogical reasoning can be implemented in AI systems as exhaustive search on a predefined analogy space. Humans' analogical ability is far more sophisticated than explicit search: it adapts to different complexity levels and task domains \citep{bethell1984adaptive}, it involves goal management and selective attention in working memory \citep{carpenter1990one, primi2001complexity}, and it requires synergy between perception and cognition that works in a bidirectional and recursive way \citep{barsalou1999perceptual, hofstadter2001analogy}. These features present a huge challenge to any existing analogy-making AI system.

Our current models use only one analogy and one transformation to solve problems in the standard Raven's test. However, multiple analogies and transformations are required for problems beyond the standard test \citep{carpenter1990one,kunda2015computational} and, thus, adding methods that coordinate multiple reasoning pathways of different analogies and transformations.

Going one step further, virtually all extant computational RPM models, including ASTI+, employ a single strategy to solve every problem. However, there is ample evidence that people change strategies on Raven's problems, sometimes within a single testing session.  For example, studies have found behavioral \citep{deshon1995verbal} and neural \citep{prabhakaran1997neural} differences across test items linked to visual versus verbal problem-solving strategies, and other dimensions of strategy may exist. How do people manage these strategies and, possibly meta-cognitively, select options appropriate for problems?  And how might an intelligent agent benefit from similar flexibility during complex problem solving?

Finally, although analogies and strategies are predefined in this research, there is the question of how humans learn such strategies, which, to our knowledge, no AI systems have accomplished for the Raven's test \citep{hernandez2016computer}. Even RPM models that use learning still require the system designer to define the function to be maximized.  Research in program induction may provide one path to tackle this thorny question \citep{schmid2011inductive}, including how strategies might be learned in the first place and adapted to new problems.

\begin{acknowledgements} 
\noindent
We would like to thank Ashok Goel for contributions to earlier phases of this research, as well as the editor for his helpful feedback.  The work was supported in part by NSF Award \#1730044.
\end{acknowledgements} 

\vspace{-0.25in}

{\parindent -10pt\leftskip 10pt\noindent
\bibliographystyle{cogsysapa}
\bibliography{main}

\begin{thebibliography}{33}
\expandafter\ifx\csname natexlab\endcsname\relax\def\natexlab#1{#1}\fi
\expandafter\ifx\csname url\endcsname\relax
  \def\url#1{{\path{\sloppy #1}}}\fi
\expandafter\ifx\csname urlprefix\endcsname\relax\def\urlprefix{From }\fi

\bibitem[{Barrett et~al.(2018)Barrett, Hill, Santoro, Morcos, \&
  Lillicrap}]{santoro2018measuring}
Barrett, D., Hill, F., Santoro, A., Morcos, A., \& Lillicrap, T. (2018).
\newblock Measuring abstract reasoning in neural networks.
\newblock {\em Proceedings of the Thirty-fifth International Conference on
  Machine Learning\/} (pp. 511--520). Cambridge, MA: PMLR.

\bibitem[{Barsalou et~al.(1999)}]{barsalou1999perceptual}
Barsalou, L.~W., et~al. (1999).
\newblock Perceptual symbol systems.
\newblock {\em Behavioral and Brain Sciences\/}, {\em 22\/}, 577--660.

\bibitem[{Benny et~al.(2021)Benny, Pekar, \& Wolf}]{benny2021scale}
Benny, Y., Pekar, N., \& Wolf, L. (2021).
\newblock Scale-localized abstract reasoning.
\newblock {\em Proceedings of the IEEE/CVF Conference on Computer Vision and
  Pattern Recognition\/} (pp. 12557--12565). Los Alamitos, CA: IEEE Computer
  Socienty.

\bibitem[{Bethell-Fox et~al.(1984)Bethell-Fox, Lohman, \&
  Snow}]{bethell1984adaptive}
Bethell-Fox, C.~E., Lohman, D.~F., \& Snow, R.~E. (1984).
\newblock Adaptive reasoning: Componential and eye movement analysis of
  geometric analogy performance.
\newblock {\em Intelligence\/}, {\em 8\/}, 205--238.

\bibitem[{Carpenter et~al.(1990)Carpenter, Just, \& Shell}]{carpenter1990one}
Carpenter, P.~A., Just, M.~A., \& Shell, P. (1990).
\newblock What one intelligence test measures: A theoretical account of the
  processing in the {Raven Progressive Matrices} test.
\newblock {\em Psychological Review\/}, {\em 97\/}, 404.

\bibitem[{Cirillo \& Str{\"o}m(2010)}]{cirillo2010anthropomorphic}
Cirillo, S., \& Str{\"o}m, V. (2010).
\newblock {\em An anthropomorphic solver for {Raven's Progressive Matrices}\/}.
\newblock Technical report, Department of Applied Information Technology,
  Chalmers University of Technology, G{\"o}teborg, Sweden.

\bibitem[{DeShon et~al.(1995)DeShon, Chan, \& Weissbein}]{deshon1995verbal}
DeShon, R.~P., Chan, D., \& Weissbein, D.~A. (1995).
\newblock Verbal overshadowing effects on {Raven's Advanced Progressive
  Matrices}: Evidence for multidimensional performance determinants.
\newblock {\em Intelligence\/}, {\em 21\/}, 135--155.

\bibitem[{Evans(1964)}]{evans1964program}
Evans, T.~G. (1964).
\newblock A heuristic program to solve geometric-analogy problems.
\newblock {\em Proceedings of the Spring Joint Computer Conference\/} (pp.
  327--338). New York, NY: ACM.

\bibitem[{Hern{\'a}ndez-Orallo et~al.(2016)Hern{\'a}ndez-Orallo,
  Mart{\'\i}nez-Plumed, Schmid, Siebers, \& Dowe}]{hernandez2016computer}
Hern{\'a}ndez-Orallo, J., Mart{\'\i}nez-Plumed, F., Schmid, U., Siebers, M., \&
  Dowe, D.~L. (2016).
\newblock Computer models solving intelligence test problems: Progress and
  implications.
\newblock {\em Artificial Intelligence\/}, {\em 230\/}, 74--107.

\bibitem[{Hespos et~al.(2020)Hespos, Anderson, \&
  Gentner}]{hespos2020structure}
Hespos, S.~J., Anderson, E., \& Gentner, D. (2020).
\newblock Structure-mapping processes enable infants’ learning across domains
  including language.
\newblock In J.~B. Childers (Ed.), {\em Language and concept acquisition from
  infancy through childhood\/},  79--104. Cham, Switzerland: Springer.

\bibitem[{Hofstadter(2001)}]{hofstadter2001analogy}
Hofstadter, D.~R. (2001).
\newblock Analogy as the core of cognition.
\newblock In D.~Gentner, K.~J. Holyoak, \& B.~N. Kokinov (Eds.), {\em The
  analogical mind: Perspectives from cognitive science\/}, chapter~15,
  499--538. Cambridge, MA: The MIT Press.

\bibitem[{Hornke \& Habon(1986)}]{hornke1986rule}
Hornke, L.~F., \& Habon, M.~W. (1986).
\newblock Rule-based item bank construction and evaluation within the linear
  logistic framework.
\newblock {\em Applied Psychological Measurement\/}, {\em 10\/}, 369--380.

\bibitem[{Hua \& Kunda(2020)}]{hua2019modeling}
Hua, T., \& Kunda, M. (2020).
\newblock Modeling gestalt visual reasoning on {Raven's Progressive Matrices}
  using generative image inpainting techniques.
\newblock {\em Proceedings of the Eighth Annual Conference on Advances in
  Cognitive Systems\/}.

\bibitem[{Hunt(1974)}]{hunt1974quote}
Hunt, E. (1974).
\newblock Quote the raven? nevermore!
\newblock In L.~W. Gregg (Ed.), {\em Knowledge and cognition\/},  129--158.
  Mahwah, NJ: Lawrence Erlbaum Associates.

\bibitem[{Kunda(2013)}]{kunda2013visual}
Kunda, M. (2013).
\newblock {\em Visual problem solving in autism, psychometrics, and {AI}: The
  case of the {Raven's Progressive Matrices} intelligence test\/}.
\newblock Doctoral dissertation, Department of Computer Science, Georgia
  Institute of Technology, Atlanta, GA.

\bibitem[{Kunda(2015)}]{kunda2015computational}
Kunda, M. (2015).
\newblock Computational mental imagery, and visual mechanisms for maintaining a
  goal-subgoal hierarchy.
\newblock {\em Proceedings of the Third Annual Conference on Advances in
  Cognitive Systems\/}. Atlanta, GA.

\bibitem[{Kunda et~al.(2013)Kunda, McGreggor, \& Goel}]{kunda2013computational}
Kunda, M., McGreggor, K., \& Goel, A.~K. (2013).
\newblock A computational model for solving problems from the {Raven’s
  Progressive Matrices} intelligence test using iconic visual representations.
\newblock {\em Cognitive Systems Research\/}, {\em 22\/}, 47--66.

\bibitem[{Kunda et~al.(2016)Kunda, Souli{\`e}res, Rozga, \&
  Goel}]{kunda2016error}
Kunda, M., Souli{\`e}res, I., Rozga, A., \& Goel, A.~K. (2016).
\newblock Error patterns on the {Raven's Standard Progressive Matrices} test.
\newblock {\em Intelligence\/}, {\em 59\/}, 181--198.

\bibitem[{Lovett \& Forbus(2017)}]{lovett2017modeling}
Lovett, A., \& Forbus, K. (2017).
\newblock Modeling visual problem solving as analogical reasoning.
\newblock {\em Psychological Review\/}, {\em 124\/}, 60.

\bibitem[{Michelson et~al.(2019)Michelson, Palmer, Dasari, \&
  Kunda}]{michelson2019learning}
Michelson, J., Palmer, J.~H., Dasari, A., \& Kunda, M. (2019).
\newblock Learning spatially structured image transformations using planar
  neural networks.
\newblock {\em arXiv preprint arXiv:1912.01553\/}.

\bibitem[{Mulholland et~al.(1980)Mulholland, Pellegrino, \&
  Glaser}]{mulholland1980components}
Mulholland, T.~M., Pellegrino, J.~W., \& Glaser, R. (1980).
\newblock Components of geometric analogy solution.
\newblock {\em Cognitive Psychology\/}, {\em 12\/}, 252--284.

\bibitem[{Nersessian(2010)}]{nersessian2010creating}
Nersessian, N.~J. (2010).
\newblock {\em Creating scientific concepts\/}.
\newblock Cambridge, MA: The MIT Press.

\bibitem[{Prabhakaran et~al.(1997)Prabhakaran, Smith, Desmond, Glover, \&
  Gabrieli}]{prabhakaran1997neural}
Prabhakaran, V., Smith, J.~A., Desmond, J.~E., Glover, G.~H., \& Gabrieli,
  J.~D. (1997).
\newblock Neural substrates of fluid reasoning: An {fMRI} study of neocortical
  activation during performance of the {Raven's Progressive Matrices} test.
\newblock {\em Cognitive Psychology\/}, {\em 33\/}, 43--63.

\bibitem[{Prade \& Richard(2014)}]{prade2014homogenous}
Prade, H., \& Richard, G. (2014).
\newblock Homogenous and heterogeneous logical proportions.
\newblock {\em Journal of Logic and Computation\/}, {\em 1\/}, 1--52.

\bibitem[{Primi(2001)}]{primi2001complexity}
Primi, R. (2001).
\newblock Complexity of geometric inductive reasoning tasks: Contribution to
  the understanding of fluid intelligence.
\newblock {\em Intelligence\/}, {\em 30\/}, 41--70.

\bibitem[{Rasmussen \& Eliasmith(2011)}]{rasmussen2011neural}
Rasmussen, D., \& Eliasmith, C. (2011).
\newblock A neural model of rule generation in inductive reasoning.
\newblock {\em Topics in Cognitive Science\/}, {\em 3\/}, 140--153.

\bibitem[{Raven et~al.(1998)Raven, Raven, \& Court}]{raven1998manual}
Raven, J., Raven, J.~C., \& Court, J.~H. (1998).
\newblock {\em Manual for {Raven's Progressive Matrices and Vocabulary
  Scales}\/}.
\newblock San Antonio, TX: Harcourt Assessment.

\bibitem[{Schmid \& Kitzelmann(2011)}]{schmid2011inductive}
Schmid, U., \& Kitzelmann, E. (2011).
\newblock Inductive rule learning on the knowledge level.
\newblock {\em Cognitive Systems Research\/}, {\em 12\/}, 237--248.

\bibitem[{Snow(1981)}]{snow1980aptitude}
Snow, R.~E. (1981).
\newblock Aptitude processes.
\newblock {\em Conference Proceedings: Aptitude, Learning, and Instruction\/}
  (pp. 27--63). London, UK: Routledge.

\bibitem[{Spearman(1923)}]{spearman1923nature}
Spearman, C. (1923).
\newblock {\em The nature of "intelligence" and the principles of cognition\/}.
\newblock London, UK: Macmillan.

\bibitem[{Sternberg(1977)}]{sternberg1977intelligence}
Sternberg, R.~J. (1977).
\newblock {\em Intelligence, information processing, and analogical reasoning:
  The componential analysis of human abilities.\/}.
\newblock Mahwah, NJ: Lawrence Erlbaum Associates.

\bibitem[{Yang et~al.(2021)Yang, Sanyal, Michelson, Ainooson, \&
  Kunda}]{yang2021automatic}
Yang, Y., Sanyal, D., Michelson, J., Ainooson, J., \& Kunda, M. (2021).
\newblock Automatic item generation of figural analogy problems: A review and
  outlook.
\newblock {\em Proceedings of the Ninth Annual Conference on Advances in
  Cognitive Systems\/}.

\bibitem[{Zhang et~al.(2019)Zhang, Gao, Jia, Zhu, \& Zhu}]{zhang2019raven}
Zhang, C., Gao, F., Jia, B., Zhu, Y., \& Zhu, S.-C. (2019).
\newblock Raven: A dataset for relational and analogical visual reasoning.
\newblock {\em Proceedings of the IEEE/CVF Conference on Computer Vision and
  Pattern Recognition\/} (pp. 5317--5327). Los Alamitos, CA: IEEE Computer
  Socienty.

\end{thebibliography}

}


\end{document}